\documentclass[11pt]{article}
\pdfoutput=1

\usepackage[preprint]{acl}

\usepackage{times}
\usepackage{latexsym}

\usepackage[T1]{fontenc}

\usepackage[utf8]{inputenc}
\usepackage{textcomp}

\DeclareUnicodeCharacter{0307}{}

\usepackage{microtype}
\usepackage{inconsolata}
\usepackage{graphicx}
\usepackage{subcaption}
\usepackage{booktabs, multirow}
\usepackage{soul}
\usepackage{changepage,threeparttable} 
\usepackage{hyperref}
\usepackage{url}
\usepackage{mathtools}
\usepackage{enumitem}

\title{Multi-round, Chain-of-thought Post-editing for Unfaithful Summaries}

\author{Yi-Hui Lee\textsuperscript{\rm 1} ~~ Xiangci Li\textsuperscript{\rm 2}$^*$ ~~ Jessica Ouyang\textsuperscript{\rm 1}\\
  \textsuperscript{\rm 1} The University of Texas at Dallas \\
  \textsuperscript{\rm 2} Amazon Web Services \\
    \texttt{lilyyhlee30@gmail.com} \\
    \texttt{lixiangci8@gmail.com} \\
    \texttt{jessica.ouyang@utdallas.edu} \\
}

\begin{document}
\maketitle

\def\thefootnote{*}\footnotetext{~Work performed as a PhD candidate at UT Dallas.}
\def\thefootnote{\arabic{footnote}}



\begin{abstract}
Recent large language models (LLMs) have demonstrated a remarkable ability to perform natural language understanding and generation tasks. In this work, we investigate the use of LLMs for evaluating faithfulness in news summarization, finding that it achieves a strong correlation with human judgments. We further investigate LLMs' capabilities as a faithfulness post-editor, experimenting with different chain-of-thought prompts for locating and correcting factual inconsistencies between a generated summary and the source news document and are able to achieve a higher editing success rate than was reported in prior work. We perform both automated and human evaluations of the post-edited summaries, finding that prompting LLMs using chain-of-thought reasoning about factual error types is an effective faithfulness post-editing strategy, performing comparably to fine-tuned post-editing models. We also demonstrate that multiple rounds of post-editing, which has not previously been explored, can be used to gradually improve the faithfulness of summaries whose errors cannot be fully corrected in a single round.
\end{abstract}

\section{Introduction}
Factual inconsistency is an important problem in text summarization. Neural abstractive summarization systems sometimes generate statements that contradict the source document or hallucinate new statements whose truth values cannot be verified by the source document. Beginning with the work of \citet{cao2018faithful}, who discovered that up to 30\% of automatically generated abstractive summaries were \textit{unfaithful} (i.e. factually inconsistent with their source document), many researchers have investigated the problem of generating more faithful summaries \citep{10.1145/3292500.3330955, falke-etal-2019-ranking, kryscinski-etal-2019-neural, zhang-etal-2023-extractive-summarization}, with the goal of reducing the prevalence of counterfactual or hallucinated statements. Other works have focused on developing metrics to quantify faithfulness (see Section \ref{sec:rw_metrics}), which can be used to rerank candidate summaries or flag unfaithful summaries for post-editing (Section \ref{sec:rw_edit_methods}).

Most prior work on post-editing unfaithful summaries has focused on correcting factual errors relating to entities in the document \citep{cao-etal-2020-factual, dong-etal-2020-multi, zhu-etal-2021-enhancing, lee-etal-2022-factual, fabbri-etal-2022-improving} because entity errors are very common \citep{pagnoni-etal-2021-understanding}, and it is relatively straightforward to replace an incorrect entity with the correct one; little to no rewriting is needed beyond a simple substitution. 

Recently, large language models (LLMs) have demonstrated strong natural language understanding and generation abilities, producing summaries that human judges strongly prefer over fine-tuned models \citep{goyal2023news}. A natural question is whether LLMs can edit an unfaithful summary into a more faithful version. In particular, the strong generation ability of LLMs may allow them to perform the more complicated rewriting needed to correct errors related to events and their properties. While a few prior works have investigated LLM post-editing, they made use of external modules, such as evidence retrieval \citep{gao-etal-2023-rarr}, or had low editing success rates \citep{liu-etal-2023-improving}. 

\begin{figure}
\centering
\includegraphics[width=\linewidth]{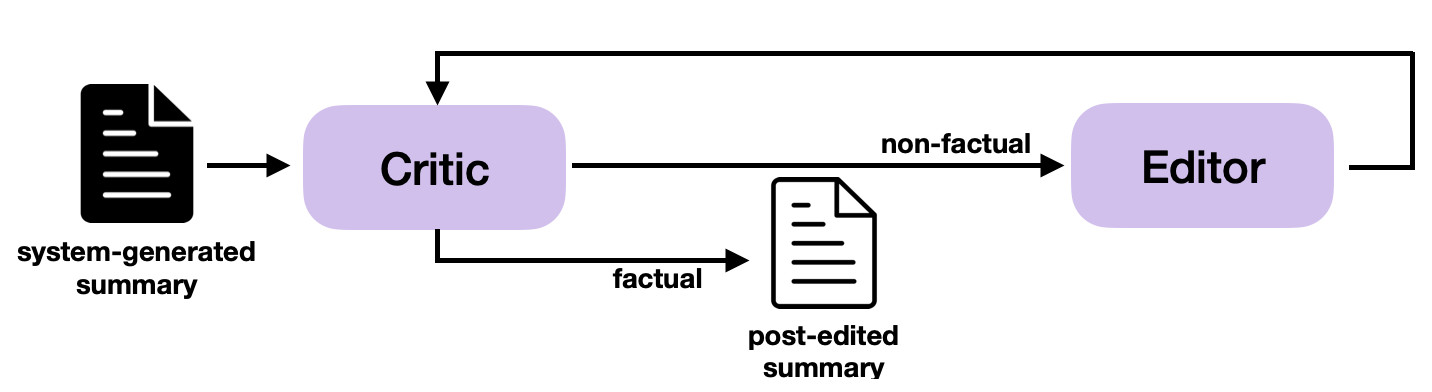}
\vspace{-0.5em}
\caption{Overview of the proposed summary post-editing approach.}
\label{fig:postedit_framework} 
\vspace{-0.5em}
\end{figure}

In this paper, we explore the use of LLMs as both faithfulness \textit{critics} and \textit{editors}: given a document-summary pair, we iteratively predict whether the summary is faithful to the document, edit if it is unfaithful, assess the faithfulness of the edited summary, and repeat until we arrive at a faithful final summary. Figure \ref{fig:postedit_framework} shows our critic-editor framework. This framework critic prompt has the opportunity to catch both remaining factual inconsistencies and any new inconsistencies that may be introduced during editing. We experiment with three different chain-of-thought \citep[CoT;][]{NEURIPS2022_9d560961} prompts for zero-shot editing of unfaithful summaries: asking the LLM to identify the summary span(s) containing factual error(s), asking it to categorize the type(s) of error(s) (e.g. an incorrect entity or time expression), and asking for both error spans and types. We evaluate the LLMs not only on the final summary editing task, but also on the auxiliary CoT tasks, and we analyze the effect of faithfulness critic and CoT task performance, as well as the error types present in the original summary, on editor performance.

Our main contributions are as follows:
\begin{itemize}
    \item We present an LLM critic-editor framework that produces more faithful summaries and achieves a higher editing success rate than prior work.

    \item Unlike prior work, which used a single post-editing step, we demonstrate that multiple rounds of editing produce more faithful final summaries, with about a 50\% improvement between the first and final editing rounds.

    \item Our experiments show that chain-of-thought reasoning about error spans and types significantly improves post-editing performance, and we find that Predicate, Entity, and Out-of-Article error types are especially important to identify for editing.
\end{itemize}

\section{Related Work}
\subsection{Faithfulness Metrics}
\label{sec:rw_metrics}
Many different types of automatic metrics have been proposed to quantify the faithfulness of a summary. \textbf{Entailment metrics} determine if the content in the generated summary is entailed by or contradictory to the input source \citep{kryscinski-etal-2020-evaluating, nie-etal-2020-adversarial, goyal-durrett-2020-evaluating}. 
\textbf{Question-answering (QA) metrics} compute a consistency score based on a QA model’s ability to answer, based the source document, questions generated from the summary \citep{wang-etal-2020-asking, durmus-etal-2020-feqa, scialom-etal-2021-questeval, fabbri-etal-2022-qafacteval}.
Finally, \textbf{large language model (LLM) metrics} \citep{gao2023humanlike, jia-etal-2023-zero, luo2023chatgpt, chen2023evaluating, fu2023gptscore, liu2023geval, lin-chen-2023-llm, cui2024dcrconsistency} prompt LLMs to predict faithfulness scores and generate explanations of the factual errors in the input summaries.

Despite the large variety of faithfulness metrics available, \citet{pagnoni-etal-2021-understanding, zhang-etal-2023-extractive} have shown that most of them have poor correlations with human judgments of factual errors in summaries. We show that our LLM critic achieves comparable or superior human correlation to three of the highest-correlation metrics --- DAE \citep{goyal-durrett-2020-evaluating}, FactCC \citep{kryscinski-etal-2020-evaluating}, and QAFactEval \citep{fabbri-etal-2022-qafacteval} --- which we use to evaluate our experimental results.

\subsection{Post-Editing for Faithfulness} 
\label{sec:rw_edit_methods}

Post-editing is a popular strategy for improving the faithfulness of a generated summary. Most approaches focus on \emph{entity errors}, where the semantic roles of entities in the summary are different from the source document. \citet{cao-etal-2020-factual, zhu-etal-2021-enhancing} train sequence-to-sequence summary correction models on synthetic data created by randomly replacing entities. \citet{dong-etal-2020-multi, lee-etal-2022-factual} target the entities directly by training models to detect incorrect entities and replace them with other candidates from the source document. \citet{fabbri-etal-2022-improving} focus on entities that are not found in the source document and propose a compression-based editing approach to remove such ``external" entities.

Most recently, \citet{gao-etal-2023-rarr} use LLMs in a two-stage, evidence-based editing approach. They generate queries from the source document to retrieve evidence from the web, compare the summary with the evidence to detect inconsistencies, and revise the summary based on the evidence. 

\citet{liu-etal-2023-improving} also experiment with LLMs, using human feedback to guide summary editing. They collect DeFacto, a dataset of (unfaithful) input summaries, human explanations of the factual errors and instructions for how to correct them, and edited summaries. They conduct experiments using zero-shot prompting with an LLM (GPT-3.5), as well as with a fine-tuned T0 model \citep{sanh2021multitask}. They find that the LLM does not perform as well as the fine-tuned model: it sometimes refuses to perform any editing and often significantly changes the summary content when it does edit. In contrast, we show that our LLM editor prompts achieve a high editing success rate and produces edited summaries that are more similar to human edits. In addition, prior post-editing approaches edit only once, while we demonstrate that multiple rounds of editing significantly improves the faithfulness of the final summary. 

Finally, \citet{madaan2024self} introduce a framework using a single LLM for output generation, feedback generation, and output refining across several tasks, including sentiment reversal, dialog response generation, code optimization/readability, mathematical reasoning, and acronym/constrained generation. However, they do not tackle summary faithfulness, and their qualitative analysis shows that their self-refining framework often suffered from erroneous self-feedback and failed to improve the original output.

\section{Methodology}

Our approach consists of two alternating steps. As Figure \ref{fig:postedit_framework} illustrates, given a source article and a system-generated input summary, we first prompt the LLM to score the faithfulness of the summary with respect to the article. If the summary is not completely faithful, we then prompt the LLM to edit the summary to be more faithful. We prompt the LLM to score the faithfulness of the new, post-edited summary, edit again if needed, and so on until the post-edited summary is predicted to be completely faithful or a maximum of five\footnote{We observed in our initial experiments that most summaries are predicted as faithful after four editing rounds, and only a handful required a fifth round.} editing rounds have occurred.

\subsection{LLM as a Faithfulness Critic}

\citet{gao2023humanlike, jia-etal-2023-zero, luo2023chatgpt} have previously investigated the use of LLM as faithfulness metrics. While faithfulness scoring with LLMs is not the focus of this work, we echo the findings of prior work that LLM faithfulness scores correlate strongly with human faithfulness judgments. Our goal is to use the LLM as a faithfulness ``critic'' that decides whether or not a system-generated summary needs post-editing, as well as how many rounds of post-editing it needs.  

Our LLM \emph{critic} scores a summary's faithfulness to its source article on a five-point Likert scale. We use in-context learning with demonstrations from the FRANK benchmark \citep{pagnoni-etal-2021-understanding}, which provides human factuality judgments for summaries generated by several different neural models on the CNN/Daily Mail \citep{nallapati-etal-2016-abstractive} and XSum \citep{narayan-etal-2018-dont} datasets.

FRANK \citep{pagnoni-etal-2021-understanding} has sentence-level binary judgments on whether there is at least one factual error in each summary sentence, and the overall summary score is the average of sentence-level scores. We bucket the overall summary scores, which range from 0 (no errors in any sentences) to 1 (at least one error in every sentence) to obtain our five-point Likert scale; for our two demonstrations, we use summaries scoring 2 and 4 on the scale, both taken from the CNN/Daily Mail partition of FRANK\footnote{The human scores in the XSum partition are at the extremes of 1 or 5, and in our initial experiments, such extreme demonstration examples resulted in the LLM critic always predicting 1.}. Despite the lack of demonstration examples from XSum, our critic performs well on that partition (Table \ref{tab:critic}). 

Figure \ref{fig:critic_prompt} shows our \textit{critic} prompt. We truncate the in-context example articles after five sentences due to our LLMs' input length limitations but do not truncate the target article.

\begin{figure}[t]
\small
\begin{center}
\setlength{\tabcolsep}{5pt}
    \begin{tabular}{p{\linewidth} }
    \hline
\textbf{System}: You are a helpful assistant that scores the faithfulness of a generated summary with respect to the summarized article. \\ \hline
\textbf{User}: Is Summary One faithful or not based on Article One? Score as 5 faithful, 4 mostly faithful, 3 neutral, 2 mostly unfaithful, or 1 unfaithful. \\
Article One: \emph{[Article One]} \\
Summary One: \emph{[Summary One]} \\
\textbf{\emph{Ranking (5, 4, 3, 2, or 1):}} \\
\textbf{Assistant}: \textbf{\emph{2}} \\ \hline
\textbf{User}: Is Summary Two faithful or not based on Article Two? Score as 5 faithful, 4 mostly faithful, 3 neutral, 2 mostly unfaithful, or 1 unfaithful. \\
Article Two: \emph{[Article Two]} \\
Summary Two: \emph{[Summary Two]} \\
\textbf{\emph{Ranking (5, 4, 3, 2, or 1):}} \\
\textbf{Assistant}: \textbf{\emph{4}} \\ \hline
\textbf{User}: Is Summary Three faithful or not based on Article Three? Score as 5 faithful, 4 mostly faithful, 3 neutral, 2 mostly unfaithful, or 1 unfaithful. \\
Article Three: \emph{[Target Article]} \\
Summary Three: \emph{[Target Summary]} \\
\textbf{\emph{Ranking (5, 4, 3, 2, or 1):}} \\ \hline
    \end{tabular}
    \vspace{-0.5em}
    \caption{Prompt format for the faithfulness critic.
    } \label{fig:critic_prompt} 
    \vspace{-1.5em}
\end{center}
\end{figure}

\subsection{LLM as a Faithfulness Editor}

Once a summary is scored as unfaithful by our LLM critic, we use the same LLM to edit a more faithful version. We design a zero-shot prompt for our LLM \textit{editor}, experimenting with two different chain-of-thought \citep{NEURIPS2022_9d560961} questions to guide editing: identifying the unfaithful spans or the error types in the input summary.

\begin{itemize}[nolistsep,noitemsep]
\item \textbf{Editor} is our baseline prompt, which asks the LLM to edit the inconsistent input summary and make it more consistent with the source document.
\item \textbf{EditorSpan} prompts the LLM to identify inconsistent spans in the input summary and then edit.
\item \textbf{EditorType} prompts the LLM to identify the error types present in the input summary and then edit. We use the error type categories of \citet{pagnoni-etal-2021-understanding} .
\item \textbf{EditorSpan+Type} prompts the LLM to identify both the inconsistent spans and error types and then edit the summary. Figure~\ref{fig:editor_prompt} shows this prompt; the prompts for the above three strategies are given in Appendix \ref{app:prompts}.
\end{itemize}

We run each LLM critic and editor in a separate session; the critic is not part of the context of the editor, and previous rounds are not part of the context of later rounds.\footnote{We experimented with running both critic and editor in a single session, but this setup resulted in dramatically lower editing success rates.}

\begin{figure}[t]
\small
\begin{center}
\setlength{\tabcolsep}{5pt}
    \begin{tabular}{p{\linewidth} }
    \hline
A summary can be inconsistent with its source article in different ways, such as \\
\textbf{Predicate Error:} The predicate in the summary is inconsistent with the source article; \\
\textbf{Entity Error}: The primary arguments (or their attributes) of the predicate are wrong; \\
\textbf{Circumstance Error}: The additional information (like location or time) specifying the circumstance around a predicate is wrong; \\
\textbf{Out of Article Error}: The summary contains information not present in the source article; \\
\textbf{Grammatical Error}: The grammar of the summary is so wrong that it becomes meaningless; \\
\textbf{Coreference Error}: A pronoun/reference with wrong or nonexisting antecedent; \\
\textbf{Discourse Link Error}: Error in how multiple summary statements are linked together in the discourse (for example temporal ordering/causal link);\\
and \textbf{Other Error}. \\ \hline
Find the span and corresponding error type(s) in the summary that is inconsistent with the source article, then edit the summary based on the inconsistent span and error types to make it more consistent with the source article in \emph{N} sentence(s):\\
Source article: \emph{[Target article]} \\
Inconsistent summary: \emph{[System-generated Target Summary]} \\ \hline
Explain your reasoning step by step and answer in the following strict format (if there are multiple inconsistent spans, give only one) \\
\textbf{\emph{Inconsistent span:}} \\
\textbf{\emph{Error types:}} \\
\textbf{\emph{Post-edited summary:}} \\
\textbf{\emph{Reasoning:}} \\ \hline
    \end{tabular}
    \vspace{-0.5em}
    \caption{Prompt format for EditorSpan+Type. Error type descriptions are from \citet{pagnoni-etal-2021-understanding}. Number of sentences $N=3$ for CNN/Daily Mail, and $N=1$ for XSum, matching the lengths of their reference summaries. 
    } \label{fig:editor_prompt}
    \vspace{-1.5em}
\end{center}
\end{figure}

\section{Faithfulness Critic Experiments}
\begin{table}[t]
\centering
\scriptsize
\begin{tabular}{lccccccc}\toprule
&\multicolumn{3}{c}{CNN/DM} &\multicolumn{3}{c}{XSum} \\\cmidrule{2-7}
 & PCC & $\rho$ &BAcc. & PCC & $\rho$ &BAcc. \\\midrule
Rouge-1 &0.27 &0.26 &- &0.18 &0.15 &- \\
Rouge-2 &0.19 &0.19 &- &0.19 &0.16 &- \\
Rouge-L &0.20 &0.20 &- &0.18 &0.14 &- \\
BertScore-R &0.36 &0.33 &- &0.14 &0.13 &- \\
\midrule
FactCC &0.49 &0.44 &0.66 &0.07 &0.07 &0.77 \\
DAE (avg.) &0.39 &0.38 &0.67 &0.24 &0.26 &\textbf{0.84} \\
QAFactEval & 0.65 &0.53 &- &0.31 &0.25 &- \\
\midrule
text-bison-001 &0.50 &0.50 &0.65 &0.35 &0.32 &0.67 \\
gemini-pro\_s &0.50 &0.48 & 0.69 & 0.39 &0.30 &0.69 \\
gemini-pro\_b & - & - & 0.51 & - & - &0.56 \\
Mixtral 8x7B\_s & 0.65 &0.59 & 0.73 & 0.39 & 0.33 &0.67 \\
Mixtral 8x7B\_b & - & - & 0.67 & - & - & 0.74\\
Llama 3.3 70B\_s &\textbf{0.66} &\textbf{0.64} &\textbf{0.75} &\textbf{0.47} &\textbf{0.38} &0.74 \\
Llama 3.3 70B\_b & - & - & 0.68 & - & - & 0.79\\
\bottomrule
\end{tabular}
\caption{\textbf{P}earson \textbf{C}orrelation \textbf{C}oefficient, Spearman's $\rho$, and balanced accuracy between automatic metrics and human judgments on the CNN/DM and XSum partitions of the FRANK dataset \citep{pagnoni-etal-2021-understanding}. s: scale 1-5 critic prompt; b: binary critic prompt.}\label{tab:critic}
\end{table}

We first conduct experiments to evaluate the performance of LLMs as faithfulness critics using the FRANK benchmark \citep{pagnoni-etal-2021-understanding}. FRANK includes summaries from both the CNN/Daily Mail and XSum datasets. Each sentence in a summary is labeled with the factual error types present in that sentence by three annotators. A sentence-level factuality score of 0 indicates there are no errors present, while a score of 1 indicates the presence of at least one labeled error; summary-level scores are obtained by averaging the sentence level scores to produce a number between 0 and 1.\footnote{Because our LLM critic scores each summary on a five-point scale, we obtain the target scores for our demonstration examples by partitioning the continuous scores from FRANK into five buckets.}

\begin{table*}[h!]
\centering
\scriptsize
\begin{tabular}{lccccccccc}\toprule
\textbf{Model} & \textbf{QAFE} & \textbf{DAE} & \textbf{FactCC} & \textbf{R1} & \textbf{R2} & \textbf{RL} & \textbf{BS-F1} & \textbf{Edit \%} \\\midrule
Original summary & 2.434 &0.795 &0.374 &\textbf{0.364} &\textbf{0.150} &\textbf{0.326} &\textbf{0.877} &- \\
\midrule
CompEdit \citep{fabbri-etal-2022-improving} & 2.675 &0.786 &0.370 &0.310 &0.115 &0.275 &0.871 &77.437 \\\midrule
EditorSpan (text-bison-001) & 3.561 &0.906 &\textbf{0.516} &0.292 &0.112 &0.261 &0.870 &76.079 \\
EditorSpan (gemini-pro) &3.482 &0.923 &0.391 &0.299 &0.110 &0.266 &0.871 &74.247 \\
EditorSpan (Mixtral 8x7B) &3.503 &\textbf{0.941} &0.403 &0.312 &0.119 &0.277 &0.872 &90.895 \\
EditorSpan (Llama 3.3 70B) &\textbf{3.642} &0.803 &0.392 &0.317 &0.121 &0.281 &0.873 &\textbf{98.575} \\

\bottomrule
\end{tabular}
\caption{Post-editing evaluation on the FRANK dataset \citep{pagnoni-etal-2021-understanding}. We compare the best-performing prompt for each of our three LLMs with CompEdit and the original (input) summary. Faithfulness metrics (QAFactEval, DAE, and FactCC) are measured against the source article; ROUGE and BERTScore are measured against the reference summary.}\label{tab:frank_editor_evaluaion}
\end{table*}

We compare the performance of our LLM critic to existing faithfulness metrics in terms of their correlation with the human scores in FRANK (Table \ref{tab:critic}). We also convert the continuous, summary-level FRANK scores to a binary ``factually correct/incorrect" label by considering any score greater than 0 to be ``incorrect." This binary label allows us to evaluate the balanced accuracy of our LLM critic; since our goal is to use the critic to decide whether or not post-editing is needed, we consider 1) binary and 2) 1 to 5 scaling critic. Binary critic predict 1 factual or 0 nonfactual, while 1 to 5 scaling predict faithfulness score of 5 to mean the input summary is already correct and does not need editing, while a score of 4 or lower means the input summary is incorrect and should be edited.\footnote{We experimented with a more relaxed stopping criterion, where editing would end upon achieving a score of 4 (mostly faithful). However, this setting resulted in lower final factuality scores and editing rates.}.

We observe that LLM faithfulness critics, regardless of the underlying LLM, correlate better with human judgments than existing metrics on the XSum dataset and are comparable to non-LLM metrics on CNN/Daily Mail. Overall, the LLM critics achieve high balanced accuracy, indicating that their predictions can reasonably be used to decide whether to run our downstream LLM editor module on a given input summary. We also see that, of the existing metrics, QAFactEval has the highest correlation with human judgment. We use QAFactEval, along with the entailment-based metrics FactCC and DAE, which have been commonly used in prior work, as our evaluation metrics in the next section.

\section{Post-Editing Experiments}
\subsection{Experimental Settings}
\paragraph{Datasets and baselines.} We evaluate our LLM editor on both the FRANK \citep{pagnoni-etal-2021-understanding} and DeFacto \citep{liu-etal-2023-improving} benchmarks. We use FRANK to analyze the effect of chain-of-thought error type prediction on post-editing and compare against the publicly available implementation of CompEdit \citep{fabbri-etal-2022-improving}. We use the XSum Hallucination Annotations of \citet{maynez-etal-2020-faithfulness} to evaluate the effect of error span prediction on post-editing performance; we use the majority vote among Maynez et al.'s three annotators.\footnote{Note that these annotations cover only the XSum partition of FRANK.}

Because the current SOTA editor, \citet{liu-etal-2023-improving}, does not report results on FRANK, we also evaluate on their DeFacto dataset. Since Liu et al.'s model is not publicly available, we use the results reported in their paper.
\vspace{-0.5em}
\paragraph{LLMs.} We evaluate our critic and editor prompts using four LLMs: Google's proprietary text-bison-001\footnote{\url{https://cloud.google.com/vertex-ai/generative-ai/docs/model-reference/text}} and gemini-pro\footnote{\url{https://cloud.google.com/vertex-ai/generative-ai/docs/model-reference/gemini}}, as well as the open-source Mixtral 8x7B using llama.cpp Q5\_K\_M quantization\footnote{\url{https://huggingface.co/TheBloke/Mixtral-8x7B-v0.1-GGUF}} and Llama 3.3 70B using ollama Q4\_K\_M quantization\footnote{\url{https://ollama.com/library/llama3.3}}. We also perform a small-scale human evaluation using GPT-4\footnote{gpt-4-0613, \url{https://platform.openai.com/docs/models/gpt-4-and-gpt-4-turbo}}, which shows the same trends as our metric evaluations and is included in Appendix \ref{app:human}. We use the same prompts for all LLMs and not perform any LLM-specific prompt engineering.

\vspace{-0.5em}
\paragraph{Metrics.} We report results using \textbf{R}OUGE \citep{lin-2004-rouge} and \textbf{B}ert\textbf{S}core-\textbf{F1} \citep{bert-score} for summarization metrics, comparing the lexical or semantic similarity of the edited summary to the reference. We use \textbf{DAE} and \textbf{FactCC} for entailment-based factuality metrics and \textbf{QA}\textbf{F}act\textbf{E}val for a QA-based factuality metric, comparing the edited summary to the source article. We also report the editing success rate: the percentage of factually inconsistent input summaries that are modified by each post-editing strategy (regardless of the factual consistency of the resulting edits).

\subsection{Experimental Results}
\label{sec:results}

\begin{table*}[t]
\centering
\scriptsize
\begin{tabular}{lcccccccccc}\toprule
\textbf{Model} & \textbf{QAFE} & \textbf{DAE} & \textbf{R1 (ref.)} & \textbf{R2 (ref.)} & \textbf{BS-F1 (ref.)} & \textbf{R1 (hum.)} & \textbf{R2 (hum.)} & \textbf{BS-F1 (hum.)} & \textbf{Edit \%} \\\midrule
Original summary &1.922 &0.704 & \textbf{0.480} & \textbf{0.253} & \textbf{0.922} & \textbf{0.759} & \textbf{0.663} & \textbf{0.954} & - \\ \midrule
CompEdit \citep{fabbri-etal-2022-improving} &2.052 &0.712 & 0.475 & 0.250 & 0.921 & 0.746 & 0.649 & 0.952 & 6.03 \\
T0 Editor \citep{liu-etal-2023-improving} &2.250 & 0.833 & 0.451 & 0.223 & - & - & - & - & - \\ \midrule
EditorSpan+Type (text-bison-001) & 2.405 &0.772 & 0.446 & 0.224 & 0.915 & 0.696 & 0.578 & 0.945 & 92.68 \\
EditorSpan+Type (gemini-pro) &2.352 &0.765 & 0.448 &0.228 &0.915 &0.682 &0.565 &0.944 & 96.53 \\
EditorSpan+Type (Mixtral 8x7B) &2.331 &0.778 & 0.451 &0.227 &0.915 &0.691 &0.573 &0.945 & 94.35\\ 
EditorSpan+Type (Llama 3.3 70B) & \textbf{2.667} & \textbf{0.856} & 0.444 & 0.221 & 0.916 & 0.709 & 0.598 & 0.947 & \textbf{98.75} \\ \midrule
*GPT-3.5 \citep{liu-etal-2023-improving} & 2.351 & 0.892 & - & - & - & 0.368 & 0.220 & - & - \\
*GPT-3.5 I \citep{liu-etal-2023-improving} & \textit{2.651} & \textit{0.910} & - & - & - & 0.722 & 0.605 & - & - \\ \midrule
Human edit  & 2.550 & 0.905 & 0.404 &0.182 &0.905 &1.000 &1.000 &1.000 & 100.00 \\ 
\bottomrule
\end{tabular}
\caption{Post-editing evaluation on the DeFacto dataset \citep{liu-etal-2023-improving}. We compare the best-performing prompt for each of our three LLMs with Liu et al.'s fine-tuned T0 Editor and human-written edits, CompEdit, and the original (input) summary. Faithfulness metrics are measured against the source article; ROUGE and BERTScore are measured against both the reference and human-edited summaries. *Liu et al.'s GPT-3.5 Editors use human error detection and editing instructions, and so are not directly comparable. Appendix Table \ref{tab:DeFacto_runtime} additionally shows the inference runtimes of models on the DeFacto dataset. }\label{tab:DeFacto_fullset_comparision}
\end{table*}

Tables \ref{tab:frank_editor_evaluaion} and \ref{tab:DeFacto_fullset_comparision} compare the best-performing prompts for each of our three LLM critic-editors with the SOTA from prior work: CompEdit on FRANK and \citet{liu-etal-2023-improving} on DeFacto. 

\begin{table*}[t]
\centering
\scriptsize
\begin{tabular}{lcccccccc}\toprule
\textbf{Prompt} & \textbf{QAFE} & \textbf{DAE} & \textbf{FactCC} & \textbf{R1} & \textbf{R2} & \textbf{RL} & \textbf{BS-F1} & \textbf{Edit \%} \\\midrule

Editor &2.549 & 0.870 &0.136 &0.203 &0.052 &0.158 &0.863 &\textbf{95.20} \\
EditorSpan & \textbf{2.707} &\textbf{0.886} &0.250 &0.221 &0.060 &0.178 &0.870 & 92.92 \\
EditorType &2.456 &0.857 &0.184 &0.209 &0.055 &0.169 &0.867 &88.10 \\
EditorSpan+Type &2.553 &0.858 &\textbf{0.281} &\textbf{0.228} &\textbf{0.066} &\textbf{0.186} &\textbf{0.874} &84.44 \\
\bottomrule
\end{tabular}
\caption{Comparison of chain-of-thought prompts with Mixtral 8x7B on the XSum partition of FRANK. Results on DeFacto, as well as other LLMs, are shown in Appendix \ref{app:results}.}\label{tab:frank_bbc_editor_evaluaion}
\end{table*}

We see that all LLMs outperform CompEdit in terms of factuality metrics and editing success rate, which is unsurprising given CompEdit only addresses entity errors and is unable to correct other error types. Since CompEdit deletes incorrect entities, we see that its ROUGE scores (against the reference summaries) are higher than those of our LLM editors, which can rewrite the summaries more significantly (see Appendix \ref{sec:preservation} for an evaluation of input summary content preservation). On the DeFacto dataset, our editors perform comparably to Liu et al's T0 Editor, which is fine-tuned on human-edited summaries. However, our LLM editors produce summaries that are much more similar to the human edits, as measured by ROUGE with respect to DeFacto's human edits. Liu et al. do not report editing success rate for these results.

For completeness, we also show two versions of Liu et al.'s GPT-3.5 Editor, but they are not directly comparable to our editors or to CompEdit; both our critic-editor loop and CompEdit perform both error \textit{detection} and correction, while Liu et al.'s GPT-3.5 Editor uses gold (human) error detection and performs only error correction. Their best-performing GPT-3.5 I model includes human-written editing instructions as part of the prompt; while they do experiment with generating the editing instructions, similar to our chain-of-thought prompting, they conclude that the LLM is not able to give good instructions, so they do not use the predicted instructions in any of their experiments.

\subsection{Impact of Chain-of-Thought Prompting}

In analyzing the effects of chain-of-thought (CoT) reasoning, we seek to answer three questions:
\begin{itemize}[nosep, noitemsep]
\item Which is more useful, error span information or error type information?
\item Is it more important to have the correct error span/type, or just to set the editor's state by attempting to predict this information?
\item Are there any specific error types that are particularly important to predict correctly?
\end{itemize}

\begin{table*}[h!]
\centering
\scriptsize
\begin{tabular}{llcccccccccc}\toprule
\textbf{Span} & \textbf{Type} & \textbf{QAFE} & \textbf{DAE} & \textbf{FactCC} & \textbf{BS-F1} & \textbf{R1} & \textbf{R2} & \textbf{RL} & \textbf{Edit \%} & \textbf{Span RL} & \textbf{Type F1} \\\midrule
- & - &2.549 &0.870 &0.136 &0.863 &0.203 &0.052 &0.158 &95.20 &- &- \\\midrule
CoT & - &2.707 &\textbf{0.886} &\textbf{0.250} &\textbf{0.870} &\textbf{0.221} &\textbf{0.060} &\textbf{0.178} &92.92 & \textbf{0.314} &- \\
Gold &- &\textbf{2.754} &0.880 &0.212 &\textbf{0.870} &\textbf{0.221} &0.058 &0.175 &\textbf{94.10} & \textit{1.000} &- \\\midrule
- & CoT &2.456 &0.857 &0.184 &0.867 &0.209 &0.055 &0.169 &89.00 & - & 0.653 \\
- & Gold &\textbf{2.767} &\textbf{0.873} &\textbf{0.252} &\textbf{0.873} &\textbf{0.220} &\textbf{0.061} &\textbf{0.180} &\textbf{93.36} & - & \textit{1.000} \\\midrule
CoT & CoT &2.553 &0.858 &\textbf{0.281} &0.874 &0.228 &0.066 &0.186 &84.44 & 0.235 & \textbf{0.685} \\
Gold & CoT &2.221 &0.815 &0.248 &0.874 &\textbf{0.235} &\textbf{0.069} &\textbf{0.193} &77.90 & \textit{1.000} & 0.667 \\
CoT & Gold &2.230 &0.812 &0.253 &\textbf{0.875} &0.233 &0.068 &0.192 &75.98 & 0.236 & \textit{1.000} \\
Gold & Gold &\textbf{2.686} &\textbf{0.867} &0.267 &0.873 &0.224 &0.062 &0.181 &\textbf{89.49} & \textit{1.000} & \textit{1.000} \\
\bottomrule
\end{tabular}
\caption{Comparison of chain-of-thought and gold error spans/types using Mixtral 8x7B. We evaluate CoT predictions using the \textbf{R}OUGE-\textbf{L} between the predicted error span and the gold span, as well as the macro-average \textbf{F1} of the predicted error types.}\label{tab:upper_bound_bbc}
\end{table*}

\begin{table*}[t]\centering
\scriptsize
\begin{tabular}{l cc c cc c cc}\toprule
 & \multicolumn{2}{c}{\textbf{PredE ($n=295$})} && \multicolumn{2}{c}{\textbf{EntE ($n=472$})} && \multicolumn{2}{c}{\textbf{OutE ($n=577$)}} \\
\cline{2-3}
\cline{5-6}
\cline{8-9}
\textbf{Metric} & \textbf{Span} & \textbf{Span+Type} && \textbf{Span} & \textbf{Span+Type} && \textbf{Span} & \textbf{Span+Type} \\
\midrule
QAFE & 3.206 & \textbf{3.303} && 3.232 & \textbf{3.467} && 2.896 & \textbf{3.049} \\
DAE & 0.940 & \textbf{0.950} && 0.942 & \textbf{0.950} && 0.918 & \textbf{0.926} \\
FactCC & 0.225 & \textbf{0.373} && 0.222 & \textbf{0.403} && 0.191 & \textbf{0.278} \\
\bottomrule
\end{tabular}
\caption{Comparison of EditorSpan and EditorSpan+Type by correctly predicted error type.}\label{tab:LLM_error_type_analysis}
\end{table*}

\paragraph{Error spans are more useful than error types.} 

As we saw in the overall results, CoT reasoning about both error spans and types generally performs the best. Table \ref{tab:frank_bbc_editor_evaluaion} breaks down the performance of our CoT prompts for Mixtral 8x7B. We show these results and perform our later analysis on the XSum partition of the FRANK dataset in order to take advantage of the availability of gold error type and span annotations.

Unsurprisingly, we find that the editing success rate of the simpler Editor and EditorSpan prompts is higher, while the more complex EditorType and EditorSpan+Type prompts sometimes fail to respond and result in lower editing rates. We note that, while \citet{liu-etal-2023-improving} do not report editing success rates, they do comment that when they prompt GPT-3.5 to generate editing instructions --- an even more complex task --- it fails to do so 23.9\% of the time. Overall, the difference between EditorSpan and EditorSpan+Type is relatively small, while Editor and EditorType lag behind, suggesting that \emph{error span information is the most helpful for post-editing}.

\paragraph{Chain-of-thought reasoning is more important than correctness for error spans.}

A natural question is whether an incorrect CoT error span/type prediction will have a cascading effect on editing performance. We investigate by directly adding the gold error spans/types to the editing prompt instead of using CoT reasoning; we use the human-annotated error spans from \citep{maynez-etal-2020-faithfulness} and error types from \citep{pagnoni-etal-2021-understanding} (see Appendix \ref{app:prompts} for these prompts).

We can see in Table \ref{tab:upper_bound_bbc} that, unsurprisingly, our editing success rate improves when we use gold error spans/types, since the prompt is simpler than the CoT version. Interestingly, while editing with gold error types outperforms CoT error types, this is not the case for error spans. We can clearly see that the CoT predicted spans are actually not very good, with low ROUGE scores compared to the gold spans, yet the edited summaries are still better than those generated using gold spans. This finding suggests that \textit{updating the LLM's state via CoT reasoning is more important than correctly identifying the exact error span.}

\paragraph{Predicate, Entity, and Out-of-Article error types are the most important to identify.}

We have seen that error span information is overall more useful for post-editing than error type information, and that the EditorSpan and EditorSpan+Type prompts perform comparably. We further investigate the effect of individual error types on post-editing performance by examining cases where EditorSpan+Type outperforms EditorSpan. Are there specific error types that are important to identify via CoT reasoning, in addition to error span information?

For a given error type, we consider all input summaries containing that error type (based on the FRANK annotations). Out of these, we identify summaries where our EditorSpan+Type prompt correctly identifies the error type (i.e. the CoT prediction and the gold error type are the same) and compare the edited summary to that produced by EditorSpan. We find three error types --- Predicate, Entity, and Out-of-Article Errors --- where EditorSpan+Type outperforms EditorSpan on all factuality metrics (Table \ref{tab:LLM_error_type_analysis}). The remaining error types (Circumstance, Grammar, Coreference, and Linkage Errors) are predicted too rarely to draw conclusions.

\begin{figure*}[h!]
    \begin{subfigure}[b]{0.5\textwidth}
        \centering
        \includegraphics[width=\textwidth]{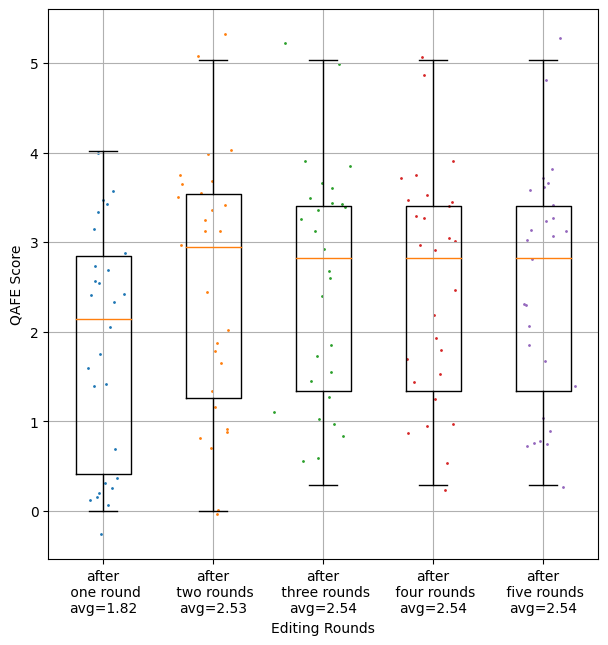}
        \vspace{-1.2em}
        \caption{EditorType}
        \label{fig:Mixtral_span_rounds}
    \end{subfigure}
    \begin{subfigure}[b]{0.5\textwidth}
        \centering
        \includegraphics[width=\textwidth]{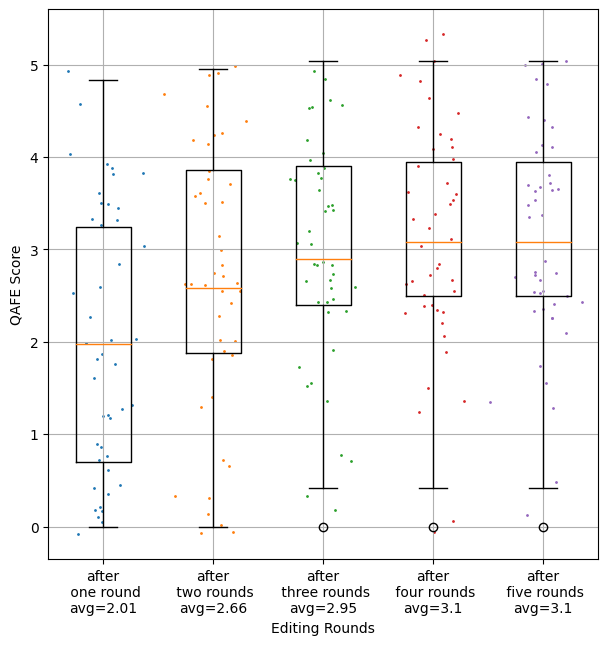}
        \vspace{-1.2em}
        \caption{EditorSpan+Type}
        \label{fig:Mixtral_both_rounds}
    \end{subfigure}
    \vspace{-1.0em}
        \caption{Comparison of summary faithfulness scores using QAFactEval across multiple rounds of editing. These results show the Mixtral 8x7B model on the FRANK dataset; other prompts and LLMs are shown in Appendix
        \ref{app:results}.}
        \label{fig:Mixtral_rounds_short}
\end{figure*}

\subsection{Impact of Multiple Editing Rounds}

Unlike prior work in post-editing, we experiment with running our critic-editor loop multiple times to perform multi-round editing. After each editor pass, we prompt our critic to rate the faithfulness of the edited summary; if any factual inconsistencies remain, or new inconsistencies have been introduced during editing, our critic should judge the edited summary to still be unfaithful, and our editing loop will run again.

Figure \ref{fig:Mixtral_rounds_short} shows a scatter plot of the QAFactEval scores of the edited summaries for each round of editing. Across all four prompt types, we see that after a single round of editing, many summaries still have low scores and are gathered near the bottom of the plot. As the number of editing rounds increases, most of those summaries improve in score and move towards the top of the plot. Most summaries exit our critic-editor loop (i.e. the critic predicts that they are now faithful) after four rounds, with only a handful of summaries requiring five rounds of editing. We do not find any correlation between the number of editing rounds an input summary requires with either its initial faithfulness score or the total number of errors in it. Appendix \ref{sec:error-analysis} shows examples of multi-round editing resulting in both correct and incorrect final summaries.

\section{Conclusion}
We have presented an LLM \textit{critic} and \textit{editor} loop for identifying and correcting unfaithful summaries with a high editing success rate. We evaluate our framework on several commonly used factuality metrics. Unlike prior work, we experiment with multiple editing rounds, guided by our faithfulness critic, finding that many summaries cannot be successfully edited in a single pass, and the faithfulness of the edited summary continues to improve over up to four rounds of editing. We also demonstrate that chain-of-thought reasoning about the factual inconsistencies in the input summary significantly improves editing performance. Locating error spans in the input summary via chain-of-thought is especially important, and we find that identifying the presence specific error types, including the common Predicate and Entity Errors, likewise helps with editing.

\section*{Limitations}
While we use LLMs as critics and editors in this work, we do not use any LLM-generated summaries. We can use only the summary systems included in the FRANK \cite{pagnoni-etal-2021-understanding} and XSum Hallucination Annotations \cite{maynez-etal-2020-faithfulness} datasets, which are accompanied by error type and error span annotations, respectively. Without such annotations, we would not be able to analyze the effects of chain-of-thought or gold error spans/types on editor performance.

\section*{Ethical Considerations}
Our framework post-edits system-generated summaries of news articles to be more faithful to the original article. However, as we show in Appendix \ref{sec:error-analysis}, our framework can still output factually incorrect final summaries. While a user may be more confident that most of our edited summaries are more faithful than the original system-generated summary, there is no guarantee that every edited summary is completely factual. 

In addition, since our framework works on news article summaries, it edits text about entities in the real world. As with any text generation system, our framework may generate misleading or incorrect statements, which may be harmful to the entities being discussed.

\bibliography{anthology,custom}

\appendix
\section{Additional Editor Prompts}
\label{app:prompts}

Figures \ref{fig:editor_prompt_plan}, \ref{fig:editor_prompt_span}, and \ref{fig:editor_prompt_type} show our Editor, EditorSpan, and EditorType LLM prompts.

\begin{figure}[h]
\small
\begin{center}
\setlength{\tabcolsep}{5pt}
    \begin{tabular}{p{\linewidth} }
    \hline
Please edit the summary to make it more consistent with the source article in \emph{N} sentence(s):\\
Source article: \emph{[Target Article]} \\
Inconsistent summary: \emph{[Input Summary]} \\ \hline
Answer in the following format \\
\textbf{\emph{Post-edited summary:}} \\ \hline
    \end{tabular}
    \vspace{-0.5em}
    \caption{Prompt format for Editor. $N=3$ for article/summary pairs from the CNN/Daily Mail partition, and $N=1$ for XSum, matching the lengths of their system-generated summaries. 
    } \label{fig:editor_prompt_plan}
    \vspace{-1.5em}
\end{center}
\end{figure}

\begin{figure}[h]
\small
\begin{center}
\setlength{\tabcolsep}{5pt}
    \begin{tabular}{p{\linewidth} }
    \hline
Find the span in the summary that is inconsistent with the source article, then edit the summary to make it more consistent with the source article in \emph{N} sentence(s):\\
Source article: \emph{[Target Article]} \\
Inconsistent summary: \emph{[Input Summary]} \\ \hline
Answer in the following format \\
\textbf{\emph{Inconsistent span:}} \\
\textbf{\emph{Post-edited summary:}} \\ \hline
    \end{tabular}
    \vspace{-0.5em}
    \caption{Prompt format for EditorSpan. $N=3$ for article/summary pairs from the CNN/Daily Mail partition, and $N=1$ for XSum, matching the lengths of their system-generated summaries. 
    } \label{fig:editor_prompt_span}
    \vspace{-1.5em}
\end{center}
\end{figure}

\begin{figure}[h]
\small
\begin{center}
\setlength{\tabcolsep}{5pt}
    \begin{tabular}{p{\linewidth} }
    \hline
A summary can be inconsistent with its source article in different ways, such as \\
\textbf{Predicate Error}: The predicate in the summary is inconsistent with the source article; \\
\textbf{Entity Error}: The primary arguments (or their attributes) of the predicate are wrong; \\
\textbf{Circumstance Error}: The additional information (like location or time) specifying the circumstance around a predicate is wrong; \\
\textbf{Out of Article Error}: The summary contains information not present in the source article; \\
\textbf{Grammatical Error}: The grammar of the summary is so wrong that it becomes meaningless; \\
\textbf{Coreference Error}: A pronoun/reference with wrong or nonexisting antecedent; \\
\textbf{Discourse Link Error}: Error in how multiple summary statements are linked together in the discourse (for example temporal ordering/causal link);\\
and \textbf{Other Error}. \\ \hline
Find the error type(s) in the summary that is inconsistent with the source article, then edit the summary based on the inconsistent error types to make it more consistent with the source article in \emph{N} sentence(s):\\
Source article: \emph{[Target Article]} \\
Inconsistent summary: \emph{[Input Summary]} \\ \hline

Explain your reasoning step by step and answer in the following strict format \\
\textbf{\emph{Error types:}} \\
\textbf{\emph{Post-edited summary:}} \\ \hline
    \end{tabular}
    \vspace{-0.5em}
    \caption{Prompt format for EditorType. Error type descriptions are from \citep{pagnoni-etal-2021-understanding}. $N=3$ for article/summary pairs from the CNN/Daily Mail partition, and $N=1$ for XSum, matching the lengths of their system-generated summaries. 
    } \label{fig:editor_prompt_type}
    \vspace{-1.5em}
\end{center}
\end{figure}

Figure \ref{fig:editor_prompt_part_upper_and_type_upper} shows the prompt EditorSpan+Type using gold error spans and types; the prompts for EditorSpan and EditorType using gold spans/types are similar.

\begin{figure}[h]
\small
\begin{center}
\setlength{\tabcolsep}{5pt}
    \begin{tabular}{p{\linewidth} }
    \hline
A summary can be inconsistent with its source article in different ways, such as \\
\textbf{Predicate Error}: The predicate in the summary is inconsistent with the source article; \\
\textbf{Entity Error}: The primary arguments (or their attributes) of the predicate are wrong; \\
\textbf{Circumstance Error}: The additional information (like location or time) specifying the circumstance around a predicate is wrong; \\
\textbf{Out of Article Error}: The summary contains information not present in the source article; \\
\textbf{Grammatical Error}: The grammar of the summary is so wrong that it becomes meaningless; \\
\textbf{Coreference Error}: A pronoun/reference with wrong or nonexisting antecedent; \\
\textbf{Discourse Link Error}: Error in how multiple summary statements are linked together in the discourse (for example temporal ordering/causal link);\\
and \textbf{Other Error}. \\ \hline
Given the span(s) and the error type(s) in the summary that are inconsistent with the source article, edit the summary based on the span and error types to make it more consistent with the source article in \emph{N} sentence(s):\\
Source article: \emph{[Target article]} \\
Inconsistent summary: \emph{[Input Summary]} \\
Inconsistent span: \emph{[Gold error span]} \\
Inconsistent error types: \emph{[Gold error types]} \\ \hline
Explain your reasoning step by step and answer in the following strict format. \\
\textbf{\emph{Post-edited summary:}} \\
\textbf{\emph{Reasoning:}} \\ \hline
    \end{tabular}
    \vspace{-0.5em}
    \caption{Prompt format for editing with given both gold error spans and gold error types instead of chain-of-thought reasoning. Error type descriptions are from \citet{pagnoni-etal-2021-understanding}. $N=3$ for article/summary pairs from the CNN/Daily Mail partition, and $N=1$ for XSum, matching the lengths of their system-generated summaries. 
    } \label{fig:editor_prompt_part_upper_and_type_upper}
    \vspace{-1.5em}
\end{center}
\end{figure}

\section{Additional Editor Results}
\label{app:results}

Table \ref{tab:frank_full_editor_evaluaion} shows the performance of all three LLMs with all four prompt variants on the full FRANK dataset, including both XSum and CNN/Daily Mail partitions.

\begin{table*}
\begin{adjustwidth}{-2.5 cm}{-2.5 cm}\centering\begin{threeparttable}[h]
\scriptsize
\begin{tabular}{llrrrrrrrrr}\toprule
\textbf{LLM} & \textbf{Prompt} & \textbf{QAFE} & \textbf{DAE} & \textbf{FactCC} & \textbf{R1} & \textbf{R2} & \textbf{RL} & \textbf{Edit \%} & \textbf{ValidEdit \%}\\\midrule
\multirow{4}{*}{text-bison-001} & Editor &3.328 & \textbf{0.914} &0.329 &0.287 &0.101 &0.254 & \textbf{81.67} & \textbf{78.69}\\
 & EditorSpan & \textbf{3.561} &0.906 &\textbf{0.516} &0.292 &0.112 &0.261 &76.08 &72.63 \\
 & EditorType &3.451 & \textbf{0.914} &0.383 & \textbf{0.301} & \textbf{0.115} & \textbf{0.270} &75.17 & 72.84\\
 & EditorSpan+Type &3.483 &0.913 & 0.476 & 0.285 &0.108 &0.253 &78.10 &73.82\\\midrule

\multirow{4}{*}{gemini-pro\_s} & Editor &3.328 &0.914 &0.329 &0.287 &0.101 &0.254 & \textbf{78.85} & \textbf{76.25}\\
 & EditorSpan &3.482 & \textbf{0.923} &0.391 & \textbf{0.299} &0.110 & \textbf{0.266} &74.25 & 73.61\\
 & EditorType &3.368 &0.903 &0.422 &0.293 &0.107 &0.261 &74.40 & 71.03\\
 & EditorSpan+Type & \textbf{3.500} &0.918 & \textbf{0.442} &0.297 & \textbf{0.111} &0.265 &71.23 & 71.59\\\midrule

 \multirow{4}{*}{gemini-pro\_b} & Editor &3.360 &\textbf{0.737} &0.394 &0.332 &0.132 &0.296 &\textbf{98.23} &\textbf{52.55} \\
 & EditorSpan &\textbf{3.361} &0.725 &0.424 &\textbf{0.337} &0.136 &0.302 &\textbf{98.23} &\textbf{52.55} \\
 & EditorType &3.182 &0.698 &0.424 &0.335 &0.136 &0.301 &88.60 &47.12 \\
 & EditorSpan+Type & 3.248 &0.699 &\textbf{0.455} &\textbf{0.337} &\textbf{0.137} &\textbf{0.303} &90.76 &48.24 \\\midrule
 
\multirow{4}{*}{Mixtral 8x7B\_s} & Editor &3.428 &0.935 &0.345 &0.301 &0.112 &0.265 & \textbf{95.87} &\textbf{77.30}\\
 & EditorSpan & \textbf{3.503} & \textbf{0.941} &0.403 & \textbf{0.313} &0.119 & \textbf{0.278} &90.90 &73.82\\
 & EditorType &3.392 &0.929 &0.371 &0.304 &0.116 &0.272 &88.94 &71.66\\
 & EditorSpan+Type &3.465 &0.931 & \textbf{0.433} &0.311 & \textbf{0.121} & \textbf{0.278} &83.83 &67.76\\\midrule
\multirow{4}{*}{Mixtral 8x7B\_b} & Editor &3.364 &0.766 &0.370 &0.314 &0.121 &0.278 &\textbf{94.81} &\textbf{66.99} \\
 & EditorSpan &\textbf{3.481} &\textbf{0.771} &0.400 &\textbf{0.320} &0.125 &0.286 &93.79 &66.50 \\
 & EditorType &3.352 &0.760 &0.415 &\textbf{0.320} &0.125 &\textbf{0.287} &92.41 &65.64 \\
 & EditorSpan+Type &3.390 &0.761 &\textbf{0.439} &0.319 &\textbf{0.126} &\textbf{0.287} &89.03 &63.07\\\midrule

 \multirow{4}{*}{Llama 3.3 70B\_s} & Editor &3.631 &\textbf{0.812} &0.382 &0.313 &0.119 &0.276 &98.73 &80.71\\
 & EditorSpan & 3.642 &0.803 &0.392 &0.317 &0.121 &0.281 &98.57 &80.71\\
 & EditorType &3.525 &0.785 &0.402 &\textbf{0.319} &\textbf{0.123} &\textbf{0.284} &86.16 &71.17\\
 & EditorSpan+Type &\textbf{3.753} &0.808 &\textbf{0.419} &0.316 &0.122 &0.281 &\textbf{98.97} &\textbf{81.13}\\\midrule
\multirow{4}{*}{Llama 3.3 70B\_b} & Editor &3.530 &0.788 &0.382 &0.324 &0.126 &0.286 &98.68 &69.78 \\
 & EditorSpan & 3.538 &0.781 &0.392 &0.325 &0.127 &0.289 &99.15 &\textbf{70.13} \\
 & EditorType &3.611 &\textbf{0.789} &\textbf{0.427} &0.325 &0.127 &0.289 &98.87 &69.92 \\
 & EditorSpan+Type &\textbf{3.631} &0.784 &0.415 &\textbf{0.326} &\textbf{0.128} &\textbf{0.290} &\textbf{99.24} &69.92 \\
 
\bottomrule
\end{tabular}
\caption{Comparison of different chain-of-thought prompts on the full FRANK dataset. s: scale 1-5 critic prompt; b: binary critic prompt.}\label{tab:frank_full_editor_evaluaion}
\end{threeparttable}\end{adjustwidth}
\end{table*}

\begin{table}[t]
\centering
\scriptsize
\begin{tabular}{lcc}\toprule
\textbf{Model} & Training Time & Inference Time \\\midrule
CompEdit \citep{fabbri-etal-2022-improving} & - & 1.3 \\
T0 Editor \citep{liu-etal-2023-improving} & 8* & -\\ \midrule
EditorSpan+Type (text-bison-001) & - & 3 \\
EditorSpan+Type (gemini-pro) & - & 3 \\
EditorSpan+Type (Mixtral 8x7B) & - & 14\\ 
EditorSpan+Type (Llama 3.3 70B) & - & 32\\
\bottomrule
\end{tabular}
\caption{Training and inference runtime (in hours) comparison on the DeFacto dataset \citep{liu-etal-2023-improving}. CompEdit, Mixtral and Llama 3.3 inference are run on one, four, and four Tesla V100S-PCIE-32GB GPUs, respectively. Bison and Gemini are run through the corresponding Google API. *Reported in \citep{liu-etal-2023-improving}.}\label{tab:DeFacto_runtime}
\end{table}

Table \ref{tab:DeFacto_runtime} shows the training and inference runtimes of our editors and baselines. CompEdit's training time is not reported in \cite{fabbri-etal-2022-improving}; we report the inference runtime of their publicly available model. \citet{liu-etal-2023-improving} do not make their model publicly available, nor do they report inference runtime; we show their reported training time for their T0 3B model trained on one 40GB GPU. We can see a clear trade-off between model size/inference time and performance for the locally-run models, CompEdit and our editors implemented with Mixtral and Llama; Llama significantly outperforms the other editors, but takes much longer to run. Our editors implemented with Bison and Gemini perform comparably to Mixtral, and their required inference time is short, even using the free tier of the Google API.

\subsection{Human Evaluation}
\label{app:human}

We perform a small-scale human evaluation of our critic-editor loop using GPT-4 as the LLM; in a pilot evaluation, we found that the GPT-4 critic had the strongest correlation with the human scores in FRANK, and our human judges preferred the output of GPT-4 over the other three LLMs. Due to the relatively high cost of GPT-4, we evaluate only 292 summaries from the FRANK dataset with human judges.

We assign two judges, all fluent in English, to evaluate each article-summary pair. Judges are shown three edited summaries, corresponding to our EditorSpan, EditorType, and EditorSpan+Type prompts, in random order and are asked to rate each summary as either factually consistent with the source article or not.

Table \ref{tab:human_consistent_num} is consistent with findings in Section \ref{sec:results} that EditorSpan+Type performs best, while EditorType performs worst.

We also collected detailed human judgments for summary Coherence, Fluency, and Relevance (to the source article) for a smaller set of 100 summaries. The scores for this smaller evaluation are shown in Table \ref{tab:human_relevance_num} and show that our EditorSpan framework with GPT-4 improves over the original input summary for every criterion.

\begin{table*}[h]\centering
\scriptsize
\begin{tabular}{lcccccc}\toprule
&\multicolumn{2}{c}{total $n=292$} &\multicolumn{2}{c}{CNN/DM $n=144$} &\multicolumn{2}{c}{XSum $n=148$} \\\cmidrule{2-7}
Model &Consistent &Inconsistent &Consistent &Inconsistent &Consistent &Inconsistent \\\midrule
EditorSpan &191 &101 &87 &57 &104 &44 \\
EditorType &160 & \textit{132} &71 & \textit{73} &89 & \textit{59} \\
EditorSpan+Type & \textbf{203} &89 &\textbf{99} &45 & \textbf{104} &44 \\
\bottomrule
\end{tabular}
\caption{Human evaluation of factually consistent/inconsistent summaries after post-editing using GPT-4.}\label{tab:human_consistent_num}
\end{table*}

\begin{table*}[h]\centering
\scriptsize
\begin{tabular}{lcccccc}\toprule
&Coherence &Consistency &Fluency &Relevance &Overall score \\\midrule
Original summary &3.215 &3.474 &3.570 &3.545 &3.573 \\
EditorSpan & \textbf{3.676} & \textbf{3.726} & \textbf{3.773} & \textbf{3.826} & \textbf{3.888} \\
\bottomrule
\end{tabular}
\caption{Human evaluation of informativeness/relevance on summaries after post-editing using GPT-4.}\label{tab:human_relevance_num}
\end{table*}

\subsection{Input Summary Preservation} \label{sec:preservation}

Because our framework involves multiple rounds of editing, we evaluate how much of the original summary content is preserved in our final edited summaries. Tables \ref{tab:frank_editor_preservation} and \ref{tab:DeFacto_fullset_preservation} show ROUGE and BERTScore measured between the original (input) summary and edited summaries from each of our LLM Editors and CompEdit on the FRANK \cite{pagnoni-etal-2021-understanding} and DeFacto \cite{liu-etal-2023-improving} datasets, respectively. We are unable to evaluate Liu et al.'s editors because their models are not publicly available.

\begin{table*}[t]
\centering
\scriptsize
\begin{tabular}{lcccccc}\toprule
\textbf{Model} & \textbf{R1 (input)} & \textbf{R2 (input)} & \textbf{RL (input)} & \textbf{BS-F1 (input)} & \textbf{Edit \%} \\\midrule
CompEdit \citep{fabbri-etal-2022-improving} & \textbf{0.770} & \textbf{0.699} & \textbf{0.767} & \textbf{0.949} &77.437 \\\midrule
EditorSpan (text-bison-001)&0.565 &0.470 &0.551 &0.919 &76.079 \\
EditorSpan (gemini-pro) &0.602 &0.510 &0.587 &0.928 &74.247 \\
EditorSpan (Mixtral 8x7B) &0.626 &0.544 &0.611 &0.933 &90.895 \\
EditorSpan (Llama 3.3 70B) &0.637	&0.553	&0.624	&0.934	&\textbf{98.575} \\

\bottomrule
\end{tabular}
\caption{Summary preservation evaluation on the FRANK dataset. We report ROUGE and BERTScore between the original (input) summary and the post-edited summary, as well as the edit success rate, for CompEdit and the best-performing prompt for each of our LLMs.}\label{tab:frank_editor_preservation}
\end{table*}

\begin{table*}[t]
\centering
\scriptsize
\begin{tabular}{lccccc}\toprule
\textbf{Model} & \textbf{R1 (input)} & \textbf{R2 (input)} & \textbf{BS-F1 (input)} & \textbf{Edit \%} \\\midrule
CompEdit \citep{fabbri-etal-2022-improving} & \textbf{0.978} & \textbf{0.974} & \textbf{0.996} & 6.03 \\ \midrule
EditorSpan+Type (text-bison-001) & 0.850 &0.805 &0.975 & 92.68 \\
EditorSpan+Type (gemini-pro) & 0.846 &0.800 &0.975 & 96.53 \\
EditorSpan+Type (Mixtral 8x7B) & 0.851 &0.806 &0.975 & 94.35 \\
EditorSpan+Type (Llama 3.3 70B) & 0.822 & 0.770 & 0.970 & \textbf{98.75} \\ \midrule
Human edit  & 0.759 &0.663 &0.954 & 100.00 \\
\bottomrule
\end{tabular}
\caption{Summary preservation evaluation on the DeFacto dataset. We report ROUGE and BERTScore between the original (input) summary and the post-edited summary, as well as the edit success rate, for CompEdit and the best-performing prompt for each of our LLMs.}\label{tab:DeFacto_fullset_preservation}
\end{table*}

We see that CompEdit very closely preserves the content of the original summary, which is unsurprising given it can only delete entities. CompEdit also achieves a relatively low editing success rate, especially on the DeFacto dataset. While our LLM Editors cause more wording changes, as shown by the lower ROUGE scores, they do so less than human editors, and the amount of change in summary meaning, as measured by BERTScore, is comparable to that of CompEdit and human editors. 

\subsection{Multi-Round Editing}\label{app:multi}

\begin{figure*}
     \centering
     \begin{subfigure}[b]{0.48\textwidth}
         \centering
         \includegraphics[width=\textwidth]{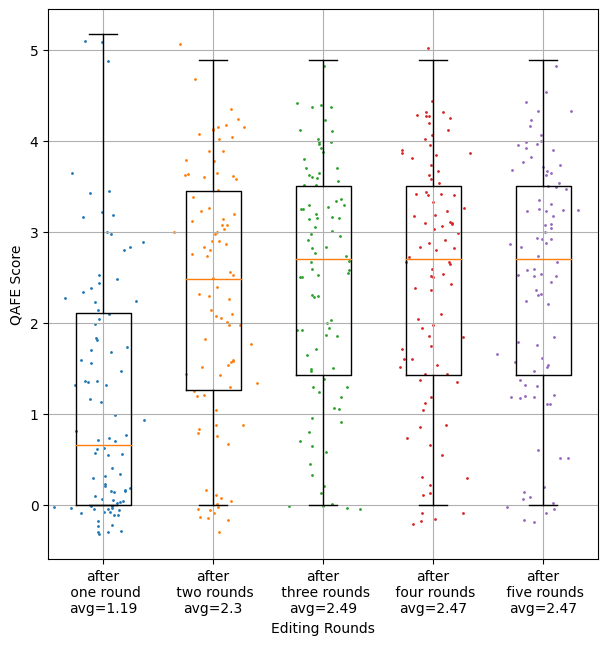}
         \vspace{-1.2em}
         \caption{Editor}
         \label{fig:Mixtral_plan_rounds}
     \end{subfigure}
     \quad
     \begin{subfigure}[b]{0.48\textwidth}
         \centering
         \includegraphics[width=\textwidth]{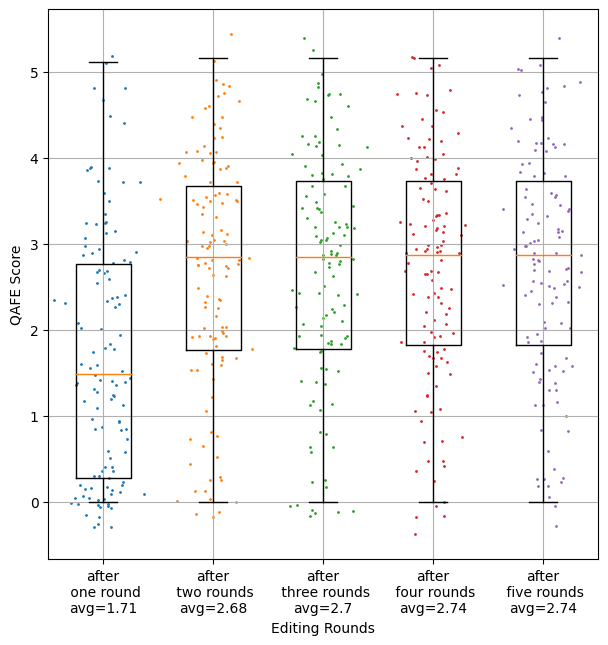}
         \vspace{-1.2em}
         \caption{EditorSpan}
         \label{fig:Mixtral_span_rounds}
     \end{subfigure}
     \quad
     \begin{subfigure}[b]{0.48\textwidth}
         \centering
         \includegraphics[width=\textwidth]{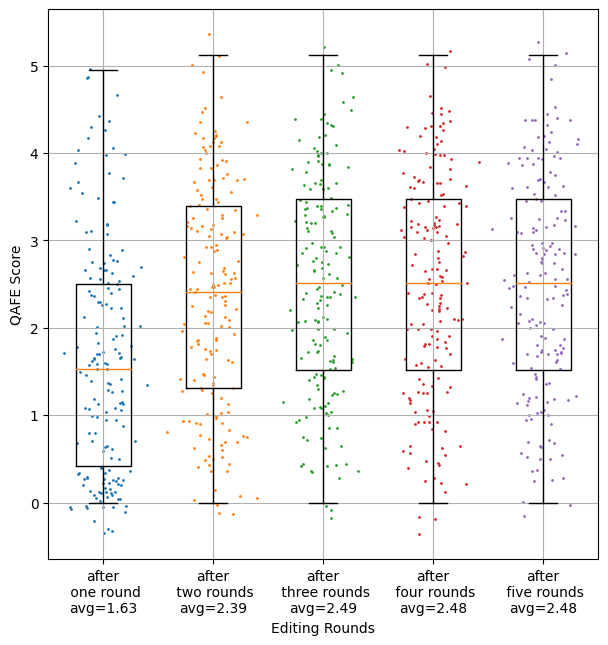}
         \vspace{-1.2em}
         \caption{EditorType}
         \label{fig:Mixtral_type_rounds}
     \end{subfigure}
     \quad
     \begin{subfigure}[b]{0.48\textwidth}
         \centering
         \includegraphics[width=\textwidth]{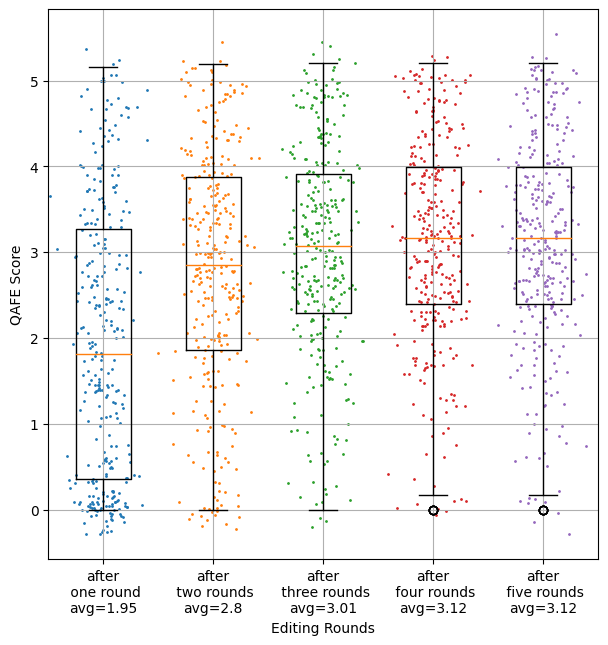}
         \vspace{-1.2em}
         \caption{EditorSpan+Type}
         \label{fig:Mixtral_both_rounds}
     \end{subfigure}
     \vspace{-1.0em}
        \caption{Full comparison of Mixtral 8x7B faithfulness scores using QAFactEval across multiple rounds of editing.}
        \label{fig:Mixtral_rounds}
\end{figure*}

\begin{figure*}
     \centering
     \begin{subfigure}[b]{0.48\textwidth}
         \centering
         \includegraphics[width=\textwidth]{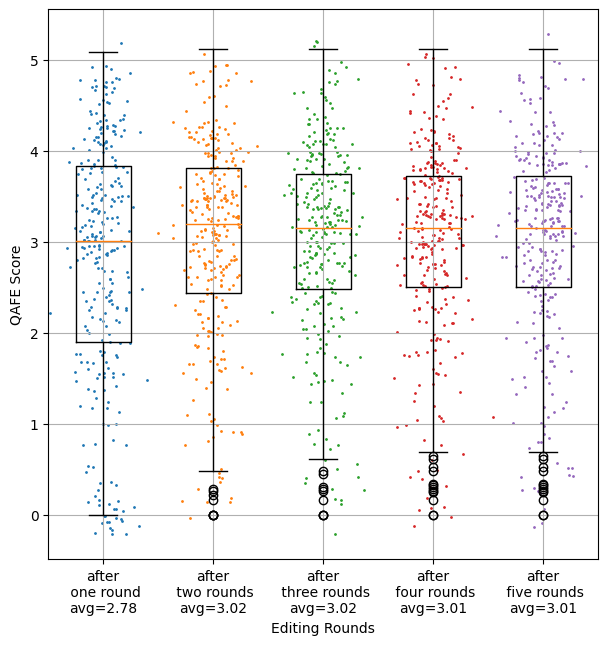}
         \vspace{-1.2em}
         \caption{Editor}
         \label{fig:Gemini_plan_rounds}
     \end{subfigure}
     \quad
     \begin{subfigure}[b]{0.48\textwidth}
         \centering
         \includegraphics[width=\textwidth]{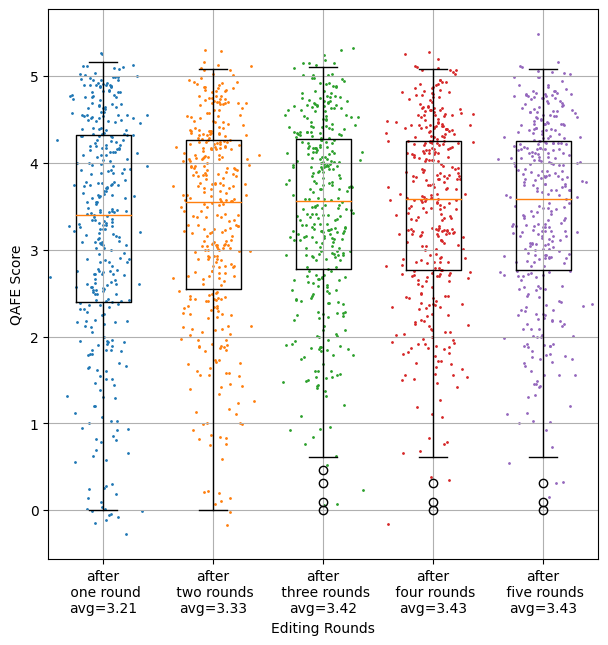}
         \vspace{-1.2em}
         \caption{EditorSpan}
         \label{fig:Gemini_span_rounds}
     \end{subfigure}
     \quad
     \begin{subfigure}[b]{0.48\textwidth}
         \centering
         \includegraphics[width=\textwidth]{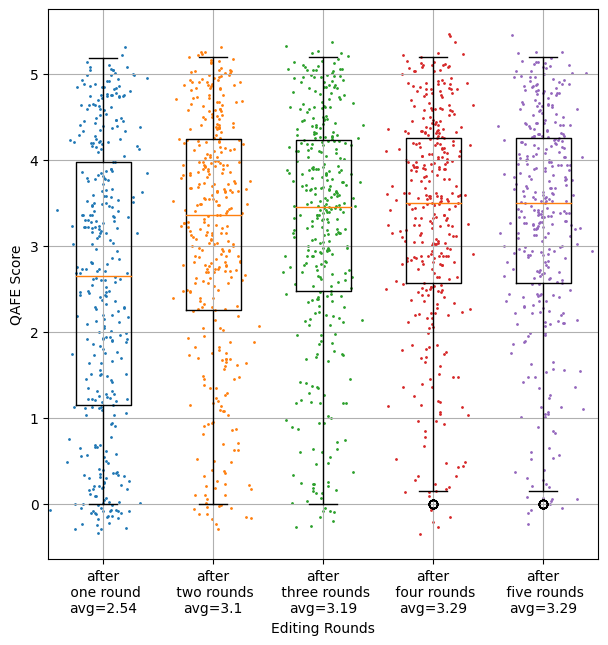}
         \vspace{-1.2em}
         \caption{EditorType}
         \label{fig:Gemini_type_rounds}
     \end{subfigure}
     \quad
     \begin{subfigure}[b]{0.48\textwidth}
         \centering
         \includegraphics[width=\textwidth]{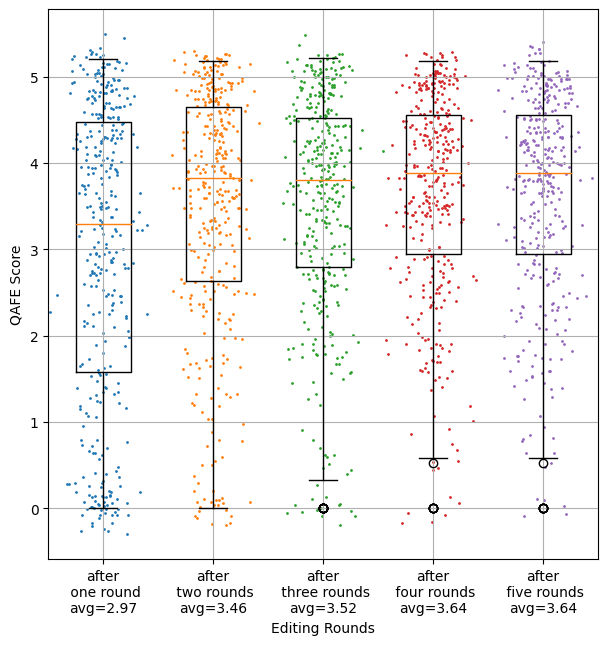}
         \vspace{-1.2em}
         \caption{EditorSpan+Type}
         \label{fig:Gemini_both_rounds}
     \end{subfigure}
     \vspace{-1.0em}
        \caption{Gemini-pro faithfulness scores using QAFactEval across multiple rounds of editing}
        \label{fig:Gemini_rounds}
\end{figure*}

\begin{figure*}
     \centering
     \begin{subfigure}[b]{0.48\textwidth}
         \centering
         \includegraphics[width=\textwidth]{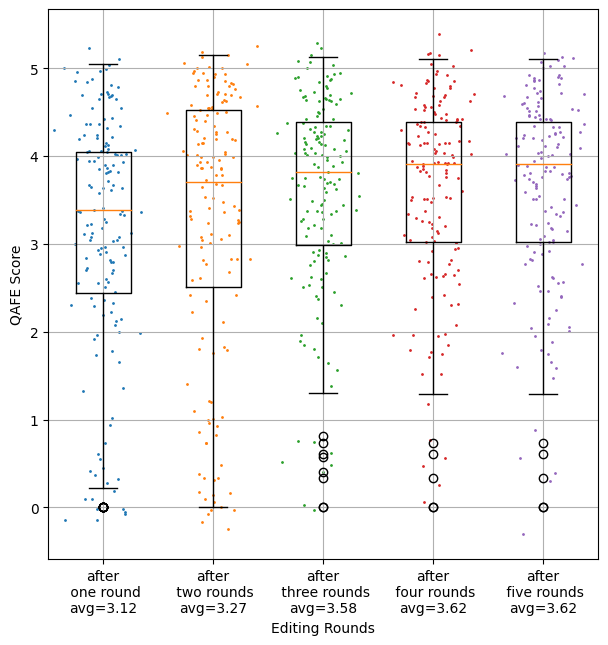}
         \vspace{-1.2em}
         \caption{Editor}
         \label{fig:Bison_plan_rounds}
     \end{subfigure}
     \quad
     \begin{subfigure}[b]{0.48\textwidth}
         \centering
         \includegraphics[width=\textwidth]{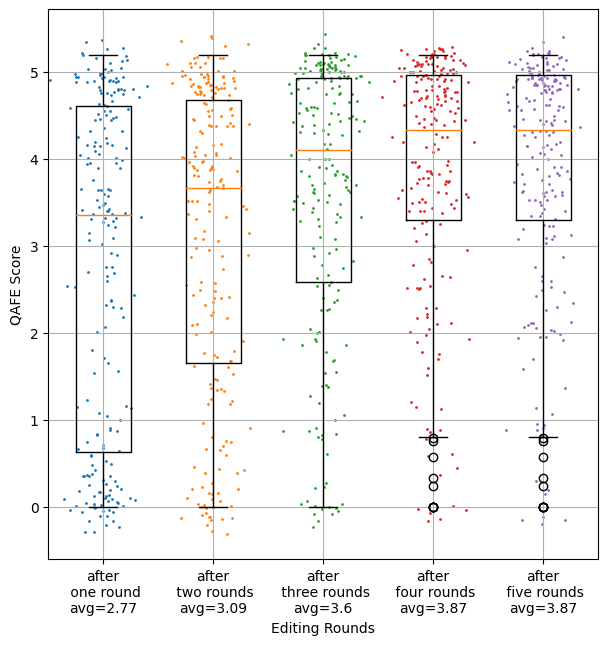}
         \vspace{-1.2em}
         \caption{EditorSpan}
         \label{fig:Bison_span_rounds}
     \end{subfigure}
     \quad
     \begin{subfigure}[b]{0.48\textwidth}
         \centering
         \includegraphics[width=\textwidth]{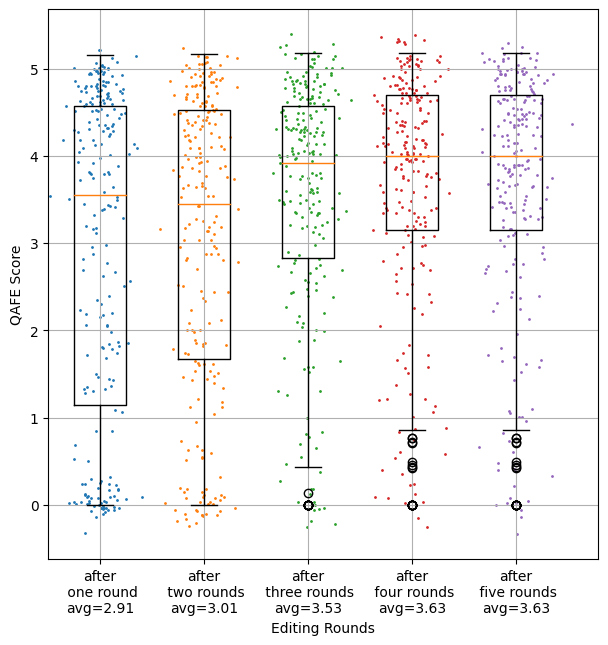}
         \vspace{-1.2em}
         \caption{EditorType}
         \label{fig:Bison_type_rounds}
     \end{subfigure}
     \quad
     \begin{subfigure}[b]{0.48\textwidth}
         \centering
         \includegraphics[width=\textwidth]{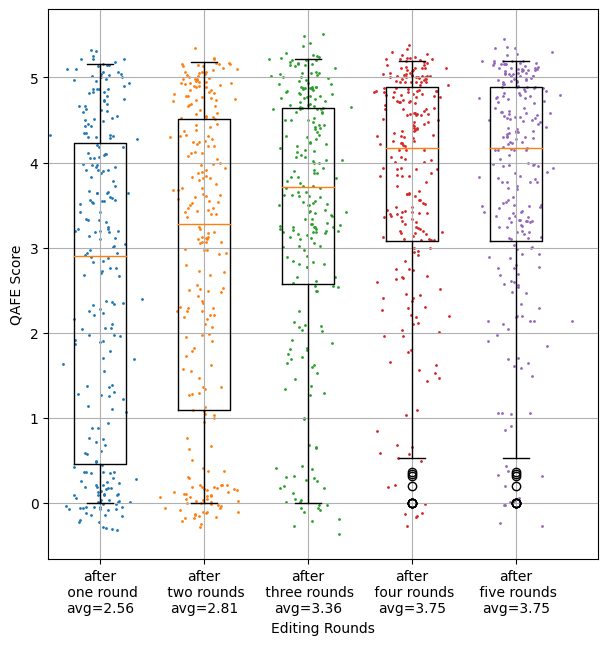}
         \vspace{-1.2em}
         \caption{EditorSpan+Type}
         \label{fig:Bison_both_rounds}
     \end{subfigure}
     \vspace{-1.0em}
        \caption{Text-bison-001 faithfulness scores using QAFactEval across multiple rounds of editing}
        \label{fig:Bison_rounds}
\end{figure*}

Figures \ref{fig:Mixtral_rounds}, \ref{fig:Gemini_rounds}, and \ref{fig:Bison_rounds} show the QAFactEval scores for edited summaries across multiple rounds of editing; we see similar trends across all three LLMs.

\begin{figure*}
     \centering
     \begin{subfigure}[b]{0.48\textwidth}
        \centering
        \includegraphics[width=\textwidth]{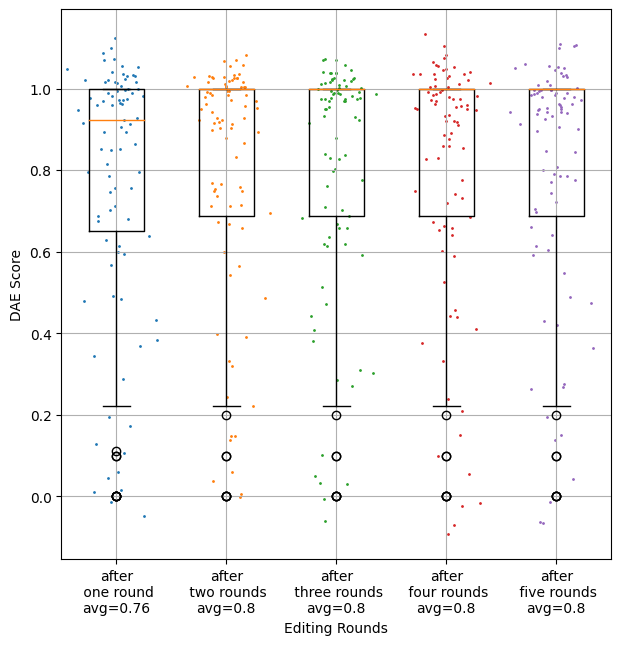}
        \vspace{-1.2em}
        \caption{Editor}
        \label{fig:Mixtral_plan_rounds}
     \end{subfigure}
     \quad
     \begin{subfigure}[b]{0.48\textwidth}
         \centering
         \includegraphics[width=\textwidth]{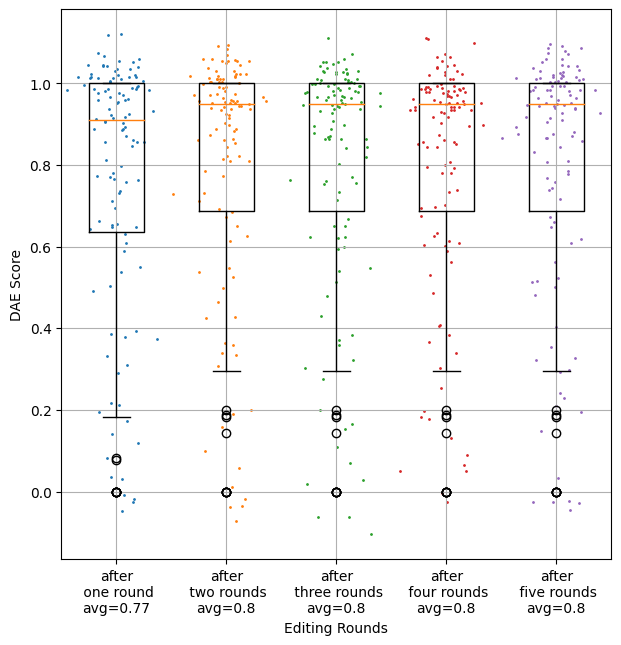}
         \vspace{-1.2em}
         \caption{EditorSpan}
         \label{fig:Mixtral_span_rounds}
     \end{subfigure}
     \quad
     \begin{subfigure}[b]{0.48\textwidth}
        \centering
        \includegraphics[width=\textwidth]{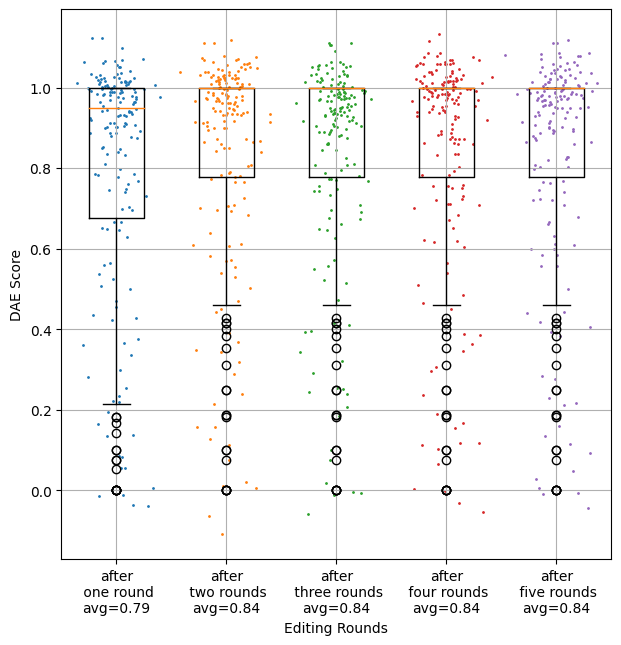}
        \vspace{-1.2em}
        \caption{EditorType}
        \label{fig:Mixtral_type_rounds}
     \end{subfigure}
     \quad
     \begin{subfigure}[b]{0.48\textwidth}
         \centering
         \includegraphics[width=\textwidth]{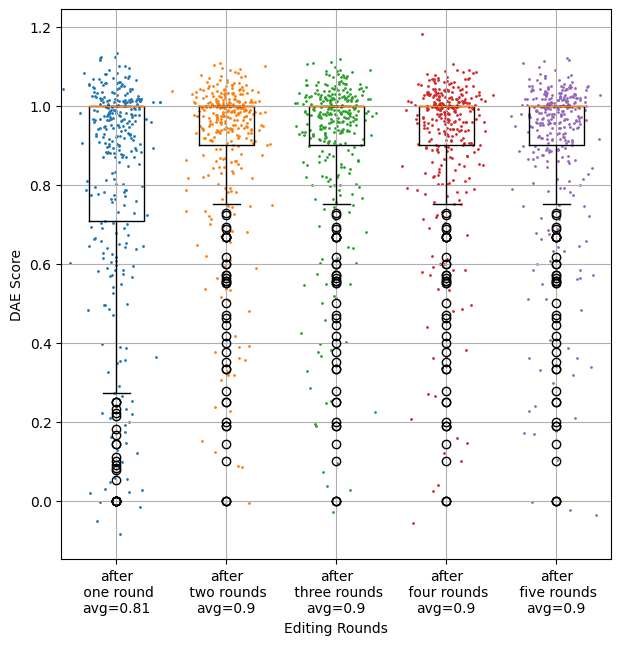}
         \vspace{-1.2em}
         \caption{EditorSpan+Type}
         \label{fig:Mixtral_both_rounds}
     \end{subfigure}
     \vspace{-1.0em}
        \caption{Full comparison of Mixtral 8x7B faithfulness scores using DAE across multiple rounds of editing.}
        \label{fig:Mixtral_rounds_DAE}
\end{figure*}

\begin{figure*}
     \centering
     \begin{subfigure}[b]{0.48\textwidth}
         \centering
         \includegraphics[width=\textwidth]{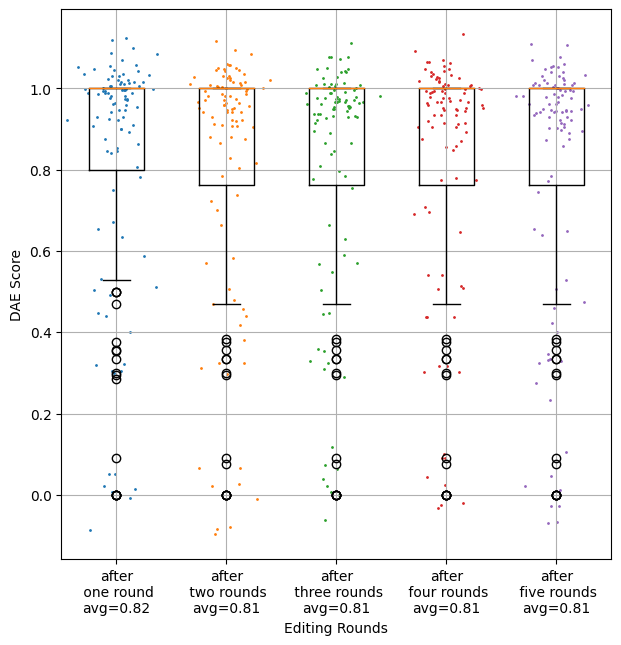}
         \vspace{-1.2em}
         \caption{Editor}
         \label{fig:Gemini_plan_rounds}
     \end{subfigure}
     \quad
     \begin{subfigure}[b]{0.48\textwidth}
         \centering
         \includegraphics[width=\textwidth]{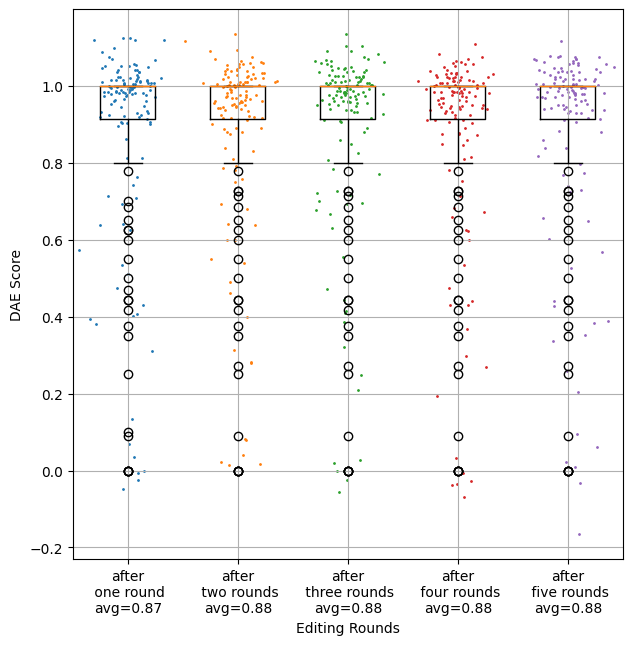}
         \vspace{-1.2em}
         \caption{EditorSpan}
         \label{fig:Gemini_span_rounds}
     \end{subfigure}
     \quad
     \begin{subfigure}[b]{0.48\textwidth}
         \centering
         \includegraphics[width=\textwidth]{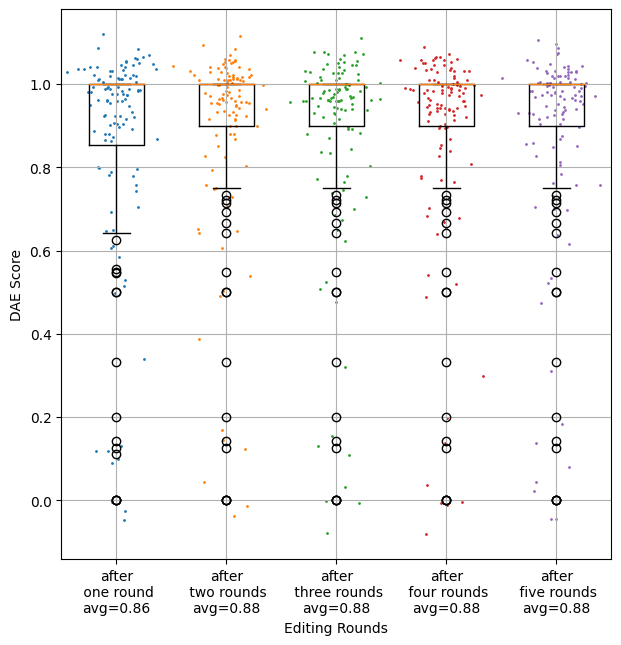}
         \vspace{-1.2em}
         \caption{EditorType}
         \label{fig:Gemini_type_rounds}
     \end{subfigure}
     \quad
     \begin{subfigure}[b]{0.48\textwidth}
         \centering
         \includegraphics[width=\textwidth]{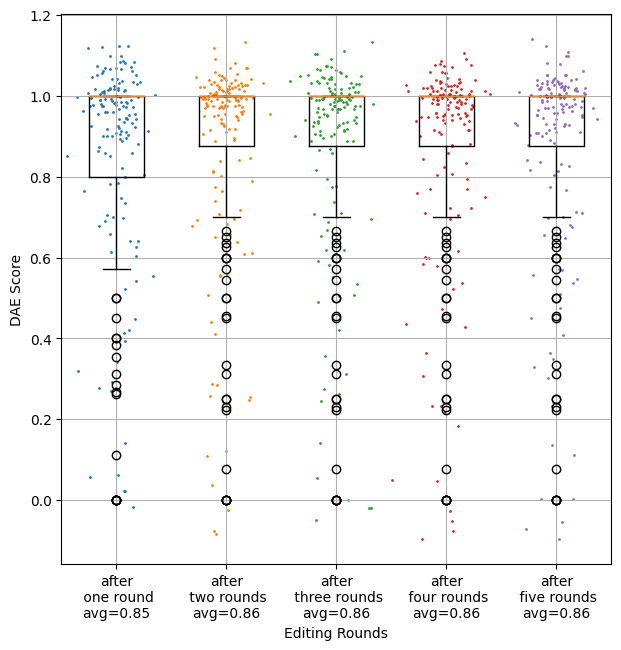}
         \vspace{-1.2em}
         \caption{EditorSpan+Type}
         \label{fig:Gemini_both_rounds}
     \end{subfigure}
     \vspace{-1.0em}
        \caption{Gemini-pro faithfulness scores using DAE across multiple rounds of editing}
        \label{fig:Gemini_rounds_DAE}
\end{figure*}

\begin{figure*}
     \centering
     \begin{subfigure}[b]{0.48\textwidth}
         \centering
         \includegraphics[width=\textwidth]{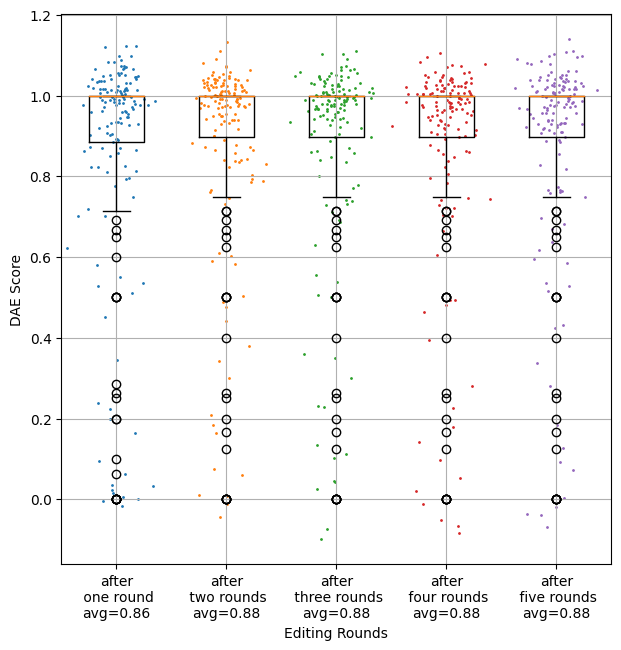}
         \vspace{-1.2em}
         \caption{Editor}
         \label{fig:Bison_plan_rounds}
     \end{subfigure}
     \quad
     \begin{subfigure}[b]{0.48\textwidth}
         \centering
         \includegraphics[width=\textwidth]{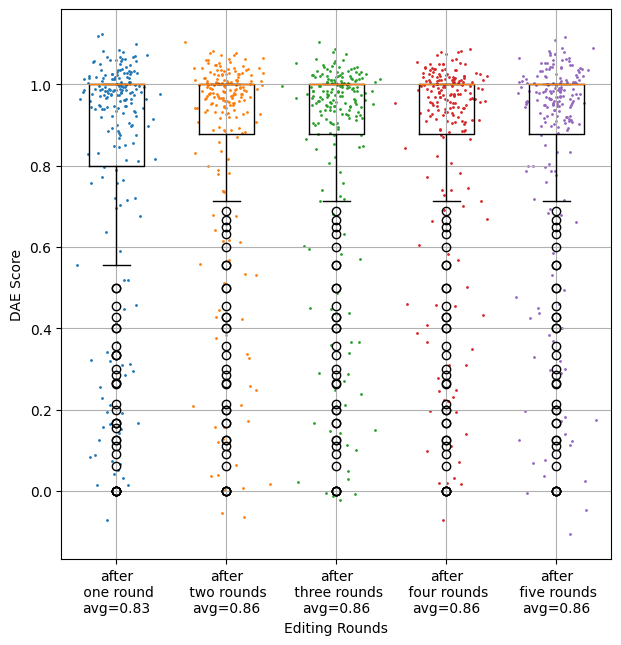}
         \vspace{-1.2em}
         \caption{EditorSpan}
         \label{fig:Bison_span_rounds}
     \end{subfigure}
     \quad
     \begin{subfigure}[b]{0.48\textwidth}
         \centering
         \includegraphics[width=\textwidth]{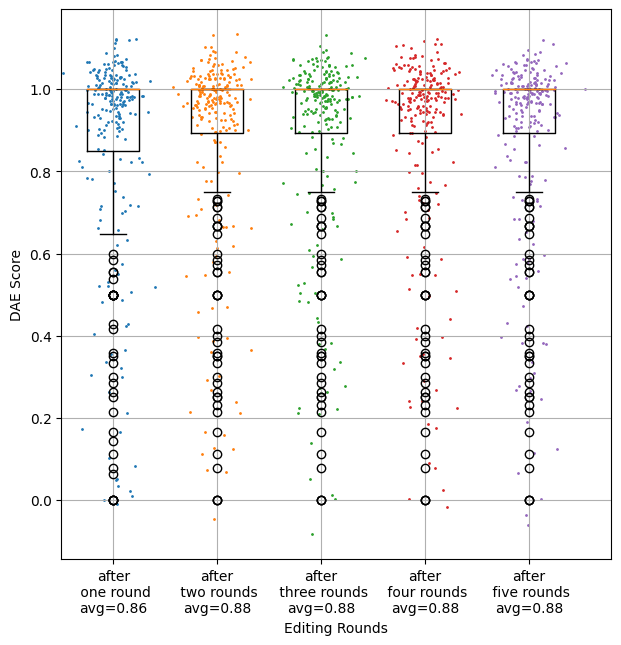}
         \vspace{-1.2em}
         \caption{EditorType}
         \label{fig:Bison_type_rounds}
     \end{subfigure}
     \quad
     \begin{subfigure}[b]{0.48\textwidth}
         \centering
         \includegraphics[width=\textwidth]{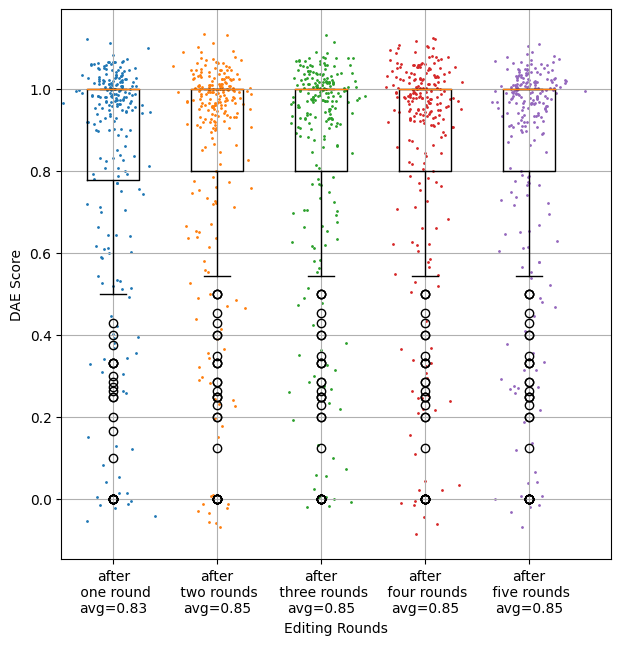}
         \vspace{-1.2em}
         \caption{EditorSpan+Type}
         \label{fig:Bison_both_rounds}
     \end{subfigure}
     \vspace{-1.0em}
        \caption{Text-bison-001 faithfulness scores using DAE across multiple rounds of editing}
        \label{fig:Bison_rounds_DAE}
\end{figure*}

Figures \ref{fig:Mixtral_rounds_DAE}, \ref{fig:Gemini_rounds_DAE}, and \ref{fig:Bison_rounds_DAE} show the DAE scores for edited summaries across multiple rounds of editing. We do not evaluate FactCC across multiple rounds because it is a binary score (0 or 1), so it is difficult to see the trends.

\subsection{Examples of Multi-Round Edited Summaries}
\label{sec:error-analysis}

We show examples of a summary improved by multi-round editing and a failure case where multi-round editing still results in an incorrect summar. Both examples are from the XSum partition of FRANK, using summaries generated with the most SOTA model available, BERTS2S. We edit the summary with our Mixtral Span Editor, which first predicts the incorrect span before editing.

\paragraph{Successful multi-round editing.}

\begin{figure*}[h]
\begin{center}
\fbox{
\begin{minipage}{.95\linewidth}
{\small
\textbf{Input summary:} \\
leyton orient have been relegated from the national league after losing to dagenham \& redbridge. \\
\vspace{-.5\baselineskip}
\hrule
\vspace{.5\baselineskip}
\textbf{Round 1:} \\

\vspace{-.5\baselineskip}
\textbf{Predicted error span:} The article mentions that Dagenham \& Redbridge won the match, not that they lost and Leyton Orient was relegated. \\
\textbf{Human-annotated error span:} leyton orient have been relegated from the national league after losing to dagenham \& redbridge \\

\vspace{-.5\baselineskip}
\textbf{Post-edited summary:} \\
Dagenham \& Redbridge kept their survival hopes alive in the National League after winning against Leyton Orient. \\
\vspace{-.5\baselineskip}
\hrule
\vspace{.5\baselineskip}
\textbf{Round 2:} \\

\vspace{-.5\baselineskip}
\textbf{Predicted error span:} The summary states "Dagenham \& Redbridge kept their survival hopes alive in the National League after winning against Leyton Orient." However, according to the source article, Dagenham and Redbridge lost the match, not won. \\

\vspace{-.5\baselineskip}
\textbf{Post-edited summary:} \\
Despite losing the match to Leyton Orient, Dagenham \& Redbridge's survival hopes in the National League remain alive due to other results. \\
\vspace{-.5\baselineskip}
\hrule
\vspace{.5\baselineskip}
\textbf{Round 3:} \\

\vspace{-.5\baselineskip}
\textbf{Predicted error span:} Despite losing the match to Leyton Orient, Dagenham \& Redbridge's survival hopes in the National League remain alive due to other results. \\

\vspace{-.5\baselineskip}
\textbf{Post-edited summary:} \\
Although Dagenham \& Redbridge lost the match and got relegated to the National League, their slim survival hopes were kept alive due to needing a victory in the article. \\
\vspace{-.5\baselineskip}
\hrule
\vspace{.5\baselineskip}
\textbf{Round 4:} \\

\vspace{-.5\baselineskip}
\textbf{Predicted error span:} their slim survival hopes were kept alive due to needing a victory. \\

\vspace{-.5\baselineskip}
\textbf{Post-edited summary:} \\
Although Dagenham \& Redbridge lost the match and got relegated to the National League, they had scored two goals that gave them brief hope of a comeback in the game.
}
\end{minipage}
}

\end{center}
    \caption{Example of successful multi-round editing.} 
    \label{fig:error-analysis-multiround}
\end{figure*}

In Figure \ref{fig:error-analysis-multiround}, the post-edited summary becomes more factual across four editing rounds; the QAFactEval score goes from 0 in the first round to 0.74 in the second round, 0.78 in the third round, and finally 2.45 in the fourth and last round, when our critic predicted that the edited summary had become faithful.

\paragraph{Unsuccessful multi-round editing and critic failure.}

\begin{figure*}[h]
\begin{center}
\fbox{
\begin{minipage}{.95\linewidth}
{\small
\textbf{Input summary:} \\
russia's lower house of parliament has approved a bill that would allow russia to issue new passports to its citizens. \\
\vspace{-.5\baselineskip}
\hrule
\vspace{.5\baselineskip}
\textbf{Round 1}: \\

\vspace{-.5\baselineskip}
\textbf{Predicted error span:} that would allow Russia to issue new passports to its citizens. \\
\textbf{Human-annotated error span:} parliament \\

\vspace{-.5\baselineskip}
\textbf{Post-edited summary:} \\
That would allow Russia to quickly issue passports to Ukrainian citizens, particularly those who are Russian-speaking and facing a `brown threat'. \\
\vspace{-.5\baselineskip}
\hrule
\vspace{.5\baselineskip}
\textbf{Round 2:} \\

\vspace{-.5\baselineskip}
\textbf{Predicted error span:} who are defenceless in the face of the `brown threat'," a reference to World War Two fascists who wore brown uniforms. \\

\vspace{-.5\baselineskip}
\textbf{Post-edited summary:} \\
That would allow Russia to quickly issue passports to Ukrainian citizens, particularly those who are Russian-speaking and reside in Crimea where pro-Russian sentiments are strong. \\
\vspace{-.5\baselineskip}
\hrule
\vspace{.5\baselineskip}
\textbf{Round 3:} \\

\vspace{-.5\baselineskip}
\textbf{Predicted error span:} That would allow Russia to quickly issue passports to Ukrainian citizens, particularly those who are Russian-speaking and reside in Crimea where pro-Russian sentiments are strong. \\

\vspace{-.5\baselineskip}
\textbf{Post-edited Summary:} \\
The bill proposed by A Just Russia party would expedite the process of issuing Russian passports to Ukrainians, especially those in the Crimea region with pro-Russian leanings, without requiring previous residence or sufficient funds and allowing them to keep their Ukrainian citizenship.
}
\end{minipage}
}
\end{center}
    \caption{Example of unsuccessful multi-round editing and the critic incorrectly predicting that an edited summary was faithful, stopping the editing process when the summary is still incorrect.} 
    \label{fig:error-analysis-critic}
\end{figure*}

In Figure \ref{fig:error-analysis-critic}, in the first round, the editor predicts and edits the wrong error span, so the edited summary was still incorrect; the editor also adds additional information about a ``brown threat", which is also incorrect. In the second round, the editor attempts to fix the “brown threat” error it previously introduced, getting further off-topic and lowering the QAFactEval score from 3.86 in the first round to 3.49. In the third and final round, the editor doubles down on the new and incorrect topic of Russian passports for Ukrainian citizens, for a final QAFactEval score of 1.53. The critic then incorrectly predicts that the rewritten summary is faithful, and the editing process stops.

\end{document}